\documentclass[twocolumn, 10pt, letterpaper]{article}
\usepackage[margin=1in]{geometry}

\usepackage{xcolor}    
\usepackage[english]{babel}
\usepackage[nottoc]{tocbibind}
\usepackage{xspace}
\usepackage{cite}
\usepackage{amsmath,amssymb,amsfonts}
\usepackage{booktabs}
\usepackage{multirow} 
\usepackage{enumitem}
\usepackage{graphicx}
\usepackage{subfig}
\usepackage{subcaption}
\usepackage{setspace}
\usepackage{float}
\usepackage{titlesec}
\usepackage{etoolbox}
\usepackage{graphicx}
\usepackage{relsize}
\usepackage{enumitem}
\usepackage{adjustbox}
\usepackage{caption}  

\usepackage{fancyhdr}

\makeatletter
\renewcommand\section{\@startsection {section}{1}{\z@}%
                                   {0.7 ex \@plus -.0 ex \@minus -.0ex}%
                                   {0.5ex \@plus.1ex}%
                                   {\normalfont\normalsize\bfseries}} 
\makeatother

\makeatletter
\renewcommand\subsection{\@startsection{subsection}{2}{\z@}%
                                     {0.4 ex \@plus -0ex \@minus -.0ex}%
                                     {0.2ex \@plus .1ex}%
                                     {\normalfont\normalsize\bfseries}} 
\makeatother

\makeatletter
\renewcommand\subsubsection{\@startsection{subsubsection}{3}{\z@}%
                                     {0.3 ex \@plus -0 ex \@minus -.0ex}%
                                     {0.2ex \@plus .1ex}%
                                     {\normalfont\normalsize\bfseries}} 
\makeatother

\setlist[itemize]{itemsep=0pt, topsep=0pt, partopsep=0pt, parsep=0pt}
\titlespacing*{\section}{0pt}{\baselineskip}{\baselineskip}
\titlespacing*{\subsection}{0pt}{\baselineskip}{\baselineskip}
\apptocmd{\thebibliography}{\setlength{\itemsep}{-0.5ex}\setlength{\parsep}{0pt}}{}{}

\titleformat{\section}
  {\normalfont\normalsize\bfseries}{\thesection}{0.5em}{}
\titleformat{\subsection}
  {\normalfont\normalsize\bfseries}{\thesection}{0.5em}{}

\newcommand{\cora}{\textbf{\texttt{Cora}}\xspace}
\newcommand{\citeseer}{\textbf{\texttt{Citeseer}}\xspace}
\newcommand{\pubmed}{\textbf{\texttt{Pubmed}}\xspace}

\newcommand{\photo}{\textbf{\texttt{Amazon-Photo}}\xspace}
\newcommand{\comp}{\textbf{\texttt{Amazon-Computers}}\xspace}

\newcommand{\cocs}{\textbf{\texttt{Coauthor-CS}}\xspace}

\DeclareMathOperator*{\argmin}{argmin}

\newcommand{\rrcl}{\texttt{GALClean}\xspace}
\newcommand{\galp}{\texttt{GALClean+}\xspace}
\newcommand{\AGE}{\texttt{AGE}\xspace}
\newcommand{\LSCALE}{\texttt{LSCALE}\xspace}
\newcommand{\Grain}{\texttt{GRAIN}\xspace}
\newcommand{\ALG}{\texttt{ALG}\xspace}
\newcommand{\Random}{\texttt{Random}\xspace}

\begin{document}

\title{Active Learning for Graphs with Noisy Structures}
\author{Hongliang Chi\thanks{Rensselaer Polytechnic Institute, chih3@rpi.edu }
\and Cong Qi\thanks{New Jersey Institute of Technology, cq5@njit.edu}
\and Suhang Wang\thanks{Penn State University, szw494@psu.edu}
\and Yao Ma\thanks{Rensselaer Polytechnic Institute, may13@rpi.edu }}
\date{}
\maketitle
\thispagestyle{fancy} 
\pagestyle{fancy}
\fancyhf{} 

\begin{abstract}
Graph Neural Networks (GNNs) have seen significant success in tasks such as node classification, largely contingent upon the availability of sufficient labeled nodes. Yet, the excessive cost of labeling large-scale graphs led to a focus on active learning on graphs, which aims for effective data selection to maximize downstream model performance. Notably, most existing methods assume reliable graph topology, while real-world scenarios often present noisy graphs. Given this, designing a successful active learning framework for noisy graphs is highly needed but challenging, as selecting data for labeling and obtaining a clean graph are two tasks naturally interdependent: selecting high-quality data requires clean graph structure while cleaning noisy graph structure requires sufficient labeled data. Considering the complexity mentioned above, we propose an active learning framework, \rrcl, which has been specifically designed to adopt an iterative approach for conducting both data selection and graph purification simultaneously with best information learned from the prior iteration. Importantly, we summarize \rrcl as an instance of the Expectation-Maximization algorithm, which provides a theoretical understanding of its design and mechanisms. This theory naturally leads to an enhanced version, \galp. Extensive experiments have demonstrated the effectiveness and robustness of our proposed method across various types and levels of noisy graphs.
\end{abstract}



\section{Introduction}\label{sec:intro}
Graph Neural Networks~\cite{kipf2016semi, wu2019simplifying} have demonstrated great potential in learning graph representation and thus facilitate the advancements of many graph-related applications including fraud detection~\cite{liu2021pick, zeng2021rlc}, recommender system \cite{gao2022graph,huang2021mixgcf}, and traffic prediction~\cite{diehl2019graph, xie2020sast}. Despite their success, GNNs typically require a large number of labeled data, especially when dealing with large-scale graphs~\cite{wu2019active}. However, it is often expensive to obtain high-quality labels. Recent efforts have been devoted to developing active learning (AL) algorithms for graphs to efficiently acquire labels with low cost~\cite{madhawa2020active, liulscale, zhang2021alg, gao2018active, hu2020graph}. Specifically, graph active learning  (GAL) aims to select a limited number of nodes for labeling, which is expected to reduce the labeling efforts by enhancing the downstream GNNs performance as much as possible. These GAL algorithms often extract and leverage the key information of nodes from both graph topology and features~\cite{cai2017active, wu2019active, liulscale}, to measure the importance of nodes and thereby perform effective node selection.  

However, the most existing GAL methods~\cite{cai2017active, wu2019active, zhang2021grain} are developed under the assumption that the underlying graph is noise-free, a condition that is rarely met in real-world applications \cite{wang2012measurement}. Moreover, as suggested in~\cite{jin2020graph}, adversarial attacks on graphs could exacerbate the situation by introducing noisy edges that connect dissimilar nodes. This structural noise can compromise GAL performance, as our \emph{Preliminary Analysis} in Section~\ref{sec:expirical} reveals. One potential solution involves cleaning the graph prior to applying GAL algorithms. However, conventional graph cleaning methods such as Pro-GNN and RS-GNN \cite{jin2020graph, dai2022towards} require labels for the cleaning process, which are not available in the active learning setting. Although unsupervised graph cleaning algorithms such as GCN-Jaccard~\cite{wu2019adversarial} do exist, typically, they can only slightly mitigate the issue of the noisy graph structure.

In this paper, we focus on addressing a significant and practical issue that has been largely overlooked: the task of conducting efficient active learning on noisy graphs. We primarily face three challenges: (i) how to accurately select valuable nodes for labeling with the presence of noisy graph structures? (ii) how to purify the noisy graph structures with limited labeled data? (iii) how to manage the complex interdependence of the first two objectives, given that the success of each is mutually dependent on the successful completion of the other? To tackle these issues, we present a novel iterative {\bf G}raph {\bf A}ctive {\bf L}earning and {\bf Clean}ing (\rrcl) framework that maximizes the synergy between node selection and graph cleaning. Specifically, to reduce the impact of structural noise on data selection, \rrcl leverages a purified graph to train a representation model to learn node representations for data selection. Moreover, a robust node selection strategy, focusing on choosing nodes that are not only valuable for the downstream task but also resilient to structural noise, is applied on the trustworthy node representations yielded prior. In parallel, those reliable representations are used to train an edge-predictor for producing a cleaner graph. This graph again reciprocates by assisting the node selection process in the next iteration with less noises. We discover that the iterative process in \rrcl can be naturally interpreted as an instance of Stochastic Expectation Maximization (Stochastic EM) algorithm~\cite{zhu2017high, balakrishnan2017statistical, papandreou2015weakly}, which provides theoretical understanding and support for \rrcl. Expanding upon this theoretical interpretation, we further introduce an enhanced framework \galp, which runs a few more iterations of EM algorithm after the labeling budget is exhausted. Extensive experiments have been done to show the effectiveness of \galp and also the importance of each key design component.

\section{Problem Definition} \label{sec:problem}
A graph is denoted as $\mathcal{G}=(\mathcal{V}, \mathcal{E})$, where $\mathcal{V}$ and $\mathcal{E}$ are the sets of nodes and edges. The connection is described as an adjacency matrix ${\bf A}\in \mathbb{R}^{N\times N}$ with $N$ denoting the number of nodes in $\mathcal{V}$. ${\bf A}_{ij}$ is the $i,j$-th element of ${\bf A}$ reflecting the strength of the connection between nodes $v_i$ and $v_j$. Each node $v_i\in \mathcal{V}$ is associated with a $d$-dimensional feature vector ${\bf x}_i \in \mathbb{R}^d$. The features for all nodes can be summarized as ${\bf X}\in \mathbb{R}^{N\times d}$. Also, each node $v_i$ has an underlying label ${\bf y}_i$. The graph $\mathcal{G}$ is assumed to contain some noises in graph structures. In particular, there is a certain proportion of edges in $\mathcal{E}$ which are heterophilous. 

The objective is to select a limited number of nodes for labeling while cleaning the graph such that a downstream GNN model trained with these labeled nodes and the cleaned graph, achieves strong performance. For this purpose, we are provided access to an oracle $\mathcal{O}$ that can supply the label of a given node within a limited budget of $B$. We are permitted to select a set of $B$ nodes from a candidate pool $\mathcal{V}_{pool} \subset \mathcal{V}$ for labeling. We denote the set of selected nodes as $\mathcal{V}_l$ and the cleaned version of the graph as $\mathcal{G}'$. This process is initialized by creating a set of labeled nodes $\mathcal{V}_{initial}$, typically containing a few nodes from each class. The process to obtain these outputs can be described as $ \mathcal{V}_l, \mathcal{G}' = \mathcal{A} \left( \mathcal{G}, {\bf X}, \mathcal{V}_{pool}, \mathcal{O}, \mathcal{V}_{initial} \right)$, 

where $\mathcal{A}$ is a graph active learning (GAL) model.

\section{Preliminary Analysis}\label{sec:expirical}
The detrimental effects of structure noises on GAL and the consequent modeling step are two folded:
\begin{itemize}[left=0em, itemsep=0.5em, labelsep=0.5em, topsep=0em, parsep=0em]
\item \textbf{Data Effect}: Current GAL methods intensively rely on graph information to identify key nodes.  The presence of noise edges can compromise the quality of the nodes selected by those methods. 
\item \textbf{Model Effect}: GNNs utilize message-passing to aggregate information from neighboring nodes on graphs. Consequently, the training and inference of downstream GNNs could be significantly distorted if noise information propagates across nodes \cite{zhu2021survey}.
\end{itemize}
To examine how the noisy graph impacts the effectiveness of existing GAL methods, we conduct empirical experiments on several recent advanced graph active learning models such as \AGE \cite{cai2017active}, \LSCALE \cite{liulscale},  \Grain \cite{zhang2021grain} and \ALG \cite{zhang2021alg}. A brief introduction to these methods can be found in Section 1 of the \textbf{supplementary file}. Specifically, we first generate a perturbed graph by randomly adding edges between nodes from different classes. The number of noisy edges added equals the number of edges in the original graph. To clearly understand the \emph{Data Effect} and \emph{Model Effect} of 
 structure noises. We run each baseline under four \underline{Noise Scenarios}. 

\begin{itemize}[left=0em, itemsep=0.2em, labelsep=0.1em, topsep=0em, parsep=0em]

\item  \textsc{Noise-Free}: Both the active learning and GCN model evaluation are performed on the clean graph.  
\item \textsc{Perturbed*}:  A perturbed graph is used for active learning but the GCN model is trained and tested on a clean graph. This setup allows us to examine the \emph{data effect} of the noisy structures.
\item \textsc{Perturbed**}: The active learning and the GCN modeling are both performed on the perturbed graph. As such, this scenario reveals the impacts of both the \emph{data effect} and the \emph{model effect}. 
\item \textsc{Pre-cleaned}: A graph pre-processing method, Jaccard-GCN \cite{wu2019adversarial}, is applied to cleanse the noisy graph and generate a pre-cleaned graph. Both the graph active learning model and the GCN modeling are conducted on this pre-cleaned graph. 
\end{itemize}

Under those cases, GAL baselines is set to choose the same number of nodes for labeling. The detailed results of this empirical investigation are shown in Table \ref{tab:emprical}. As shown in the results, it is evident that the use of a perturbed graph in either the active learning step or the GNN training and inference steps can strongly impair the models' performance. This highlights the necessity for a robust GAL model capable of selecting high-quality nodes in a noisy setting and also producing a cleaner graph to facilitate the modelling of downstream GNNs. Notably, under the \textsc{Pre-cleaned} scenario, GAL methods achieve better performance compared with the \textsc{Perturbed**} setting. However, their performances are still significantly worse than those under the \textsc{Noise-Free} scenario.

\begin{table}[h!tb]
\small\relsize{-1.2}
\caption{ Node classification performance under four different noise conditions.}
\begin{tabular}{ccccc}
\toprule
Model  &   \underline{Noise Scenario}  &   \cora & \citeseer & \pubmed \\ 
\midrule
\multirow{3}{*}{\AGE}  &    \textsc{Noise-Free}   &  77.07\% & 68.26\%   & 76.52\%  \\ 
     &   \textsc{Perturbed*} &  76.09\% & 67.21\%   & 73.60\% \\ 
   &  \textsc{Perturbed**} &  52.74\% & 45.95\%   & 54.91\%  \\ 
& \textsc{Precleaned} & 57.50\% & 51.56\% & 56.95\% \\ 
   \midrule
\multirow{3}{*}{\LSCALE}  &     \textsc{Noise-Free}   &  78.54\% & 68.79\%   & 78.39\% \\ 
      & \textsc{Perturbed*} &  76.38\% & 65.87\%  & 72.93\%  \\ 
      & \textsc{Perturbed**} &  50.31\% & 44.50\%   & 53.48\%  \\ 
      & \textsc{Precleaned} & 57.12\% & 49.67\% & 55.22\% \\ 
      \midrule
\multirow{3}{*}{\Grain}  &    \textsc{Noise-Free}  &  78.42\%  & 68.46\%   & 78.27\%  \\ 
      & \textsc{Perturbed*} &  78.31\%  & 66.36\%   & 75.97\%  \\ 
      & \textsc{Perturbed**} &  53.76\%  & 48.41\%   & 56.61\%  \\ 
      & \textsc{Precleaned} & 61.26\% & 54.80\% & 56.83\% \\ 
      \midrule
\multirow{3}{*}{\ALG} &    \textsc{Noise-Free}  &   77.68\%  & 69.44\%   & 78.66\%  \\ 
  &  \textsc{Perturbed*} &  75.91\% &  68.15\%   & 75.53\%  \\ 
  & \textsc{Perturbed**} & 52.10\% &  47.54\%   & 56.81\%  \\ 
  & \textsc{Precleaned} & 58.73\% & 53.99\% & 59.45\% \\ 
  \bottomrule
\end{tabular}
\label{tab:emprical}
\end{table}

\section{Methodology}

\begin{figure*}[htp]%
\vspace{-1cm}
\centering
\includegraphics[width=0.68\linewidth]{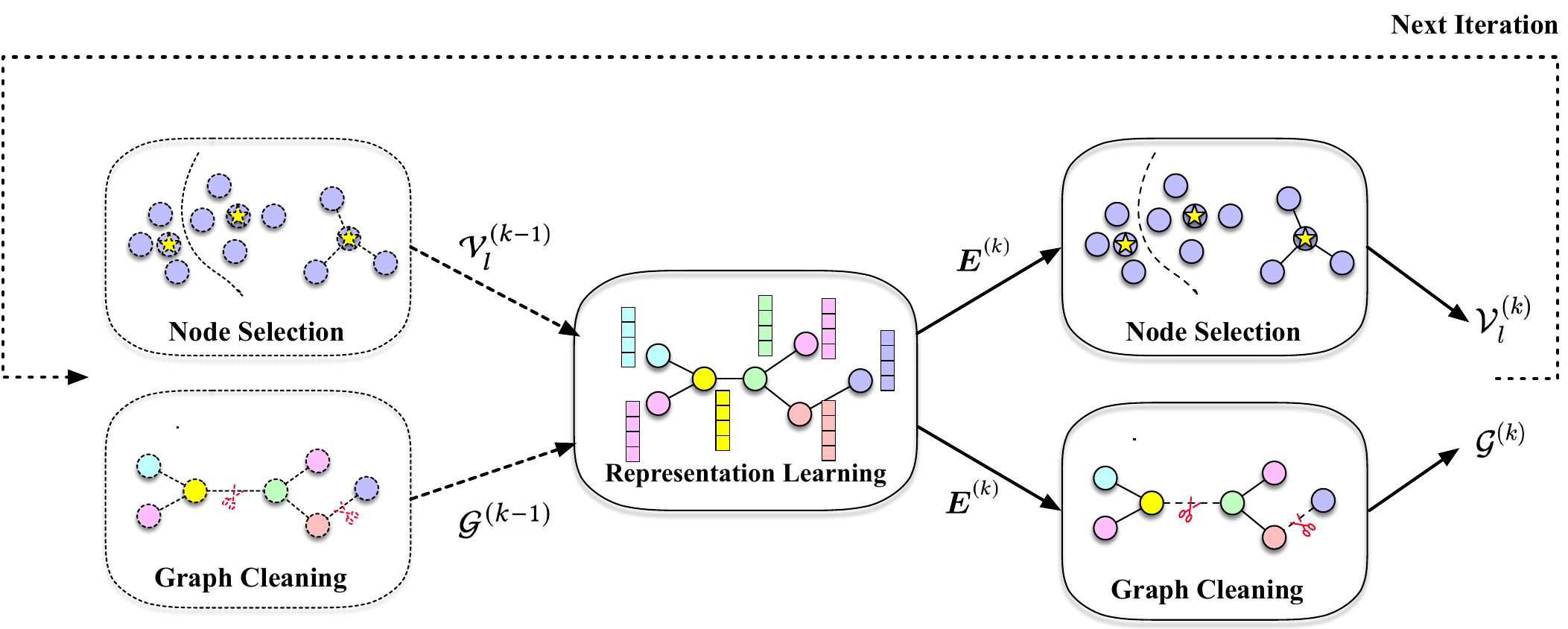}
\caption{\small{Overall framework of \rrcl. }}
\label{fig:framework}
\vspace{-0.5cm}
\end{figure*}

Given the observed {\bf Data Effect} and {\bf Model Effect}, a framework that can concurrently address graph cleaning and node selection is highly needed. Following this lead, we propose \rrcl, a solution that not only achieves both node selection and graph cleaning but also ensures they mutually help each other's effectiveness.

\subsection{\rrcl Framework}\label{sec:gcl_method}
At \rrcl, we iterate over \emph{representation learning}, \emph{node selection}, and \emph{graph cleaning} modules multiple times to gradually gather labeled nodes and purify the graph. A visual demonstration is provided at Figure~\ref{fig:framework}. In essence, both the \emph{node selection} and \emph{graph cleaning} components aim to extract more information from the graph data. This is achieved by acquiring additional labeled data and mitigating noise within the graph structures from two perspectives. The information thus obtained is utilized to learn high-quality node representations in the \emph{representation learning} component, which informs further iterations of \emph{node selection} and \emph{graph cleaning}. 

To provide an overview of \rrcl, we use the $k$-th iteration to briefly illustrate this process. Specifically, we define the labeled set and the graph after $(k-1)$-th iteration as $\mathcal{V}_l^{(k-1)}$ and $\mathcal{G}^{(k-1)}$, respectively. Then, in the $k$-th iteration, we first utilize $\mathcal{V}_l^{(k-1)}$ and $\mathcal{G}^{(k-1)}$ to learn node representations ${\bf E}^{(k)}$ by training a \emph{representation learning} model. These representations ${\bf E}^{(k)}$ incorporate the additional gained information from $(k-1)$-th iteration. They are utilized to expand the labeled set to $\mathcal{V}_l^{(k)}$ in the \emph{node selection} component and obtain a cleaner graph $\mathcal{G}^{(k)}$ in the \emph{graph cleaning} component by utilizing reliable pseudo labels derived from them. $\mathcal{V}_l^{(k)}$ and  $\mathcal{G}^{(k)}$ will then be utilized to conduct the $(k+1)$-th iteration of the process. To initialize this process, we set $\mathcal{V}_l^{(0)} = \mathcal{V}_{initial}$ and $\mathcal{G}^{(0)} = \mathcal{G}$. Note that $\mathcal{V}_{initial}$ and $\mathcal{G}$ are introduced in Section~\ref{sec:problem}. After a total of $K$ iterations, we exhaust the labeling budget and conclude with a labeled set of nodes $\mathcal{V}_l^{(K)}$ and a graph $\mathcal{G}^{(K)}$. The final set of labeled nodes, $\mathcal{V}_l^{(K)}$, is also the output of the framework, i.e., $\mathcal{V}_l = \mathcal{V}_l^{(K)}$. 
\subsubsection{Representation Learning}
The \emph{representation learning} module in \rrcl is designed to learn trustworthy node representations as the input of \emph{node selection} and \emph{graph cleaning} modules. Current GAL methods often learn node representations using GNNs with supervision from labeled data~\cite{cai2017active, gao2018active}. However, the GCN incorporates both the supervision signals and graph structural information in a coupled manner.
 For noisy graphs, a major limitation of this coupled design is that it inevitably includes structural noises, leading to undesirable node representations. To address this issue, as inspired by~\cite{hu2021graph, gasteiger2018predict}, we propose to decouple the process of capturing the graph structural information and the label information. With the set of labeled nodes $\mathcal{V}^{(k-1)}_l$ and the graph $\mathcal{G}^{(k-1)}$ produced in the $(k-1)$-th iteration, the overall objective is as follows.
{\small
\begin{align*}
    \mathcal{L} = \mathcal{L}_l (\mathcal{V}_l^{(k-1)}, {\bf E}^{(k)})  + \alpha \mathcal{L}_{g}(\mathcal{G}^{(l-1)}, {\bf E}^{(k)}),\\
{\bf E}^{(k)} = MLP_1({\bf X}; {\bf W}_1^{(k)}),
\end{align*}
}
where ${\bf E}^{(k)}$ denotes the representations produced by Multi-layer Perceptron (MLP) with the original node features ${\bf X}$ as input and ${\bf W}_1^{(k)}$ denotes all parameters of the MLP model in $k$-th iteration. The terms $\mathcal{L}_l$ and $\mathcal{L}_g$ capture label information and graph structural information respectively. The hyper-parameter $\alpha$ balances the two terms. Specifically,  $\mathcal{L}_l$ is the classification loss of MLP. 
{\small 
\begin{align*}
\mathcal{L}_{l} =\sum_{v_{i} \in \mathcal{V}_l^{(k-1)}} \ell({\bf y}_i, {\bf p}^{(k)}_i), \ \text{with }  {\bf p}^{(k)}_i = MLP_2({\bf E}_i^{(k)}; {\bf W}_2^{(k)} )
\end{align*}
}
where ${\bf p}^{(k)}_i$ denotes the vector of logits for node $v_i$ obtained by transforming ${\bf E}^{(k)}_i$ through $MLP_2$, and $\ell(\cdot)$ is the cross entropy loss. 
To capture the graph structural information in $\mathcal{G}^{(k-1)}$, we adapt the neighborhood contrastive loss~\cite{hu2021graph} to include the edge strength weights learned in the previous iteration. Specifically, the adjacency matrix ${\bf A}^{(k-1)}$ of the graph $\mathcal{G}^{(k-1)}$ contains these edge strength weights (the process to obtain ${\bf A}^{(k-1)}$ will be introduced in the \emph{graph cleaning} process in Section~\ref{sec:gclean}. In particular, $A^{(k-1)}_{ij}$ is non-zero only when node $v_i$ and $v_j$ are connected and a larger value of $A^{(k-1)}_{ij}$ indicates a higher probability that the edge between them is ``clean''. The adapted neighborhood contrastive loss is as follows. 
{\small
\begin{align*}
\mathcal{L}_{g} =- \sum_{v_{i} \in \mathcal{V}} \log \frac{\sum_{j=1}^N   {\bf A}^{(k-1)}_{ij}\exp \left(\operatorname{cos}\left({\bf E}^{(k)}_{i}, {\bf E}^{(k)}_{j}\right) / \tau\right)}{\sum_{v_m\in \mathcal{M}(v_i)}  \exp \left(\operatorname{cos}\left({\bf E}^{(k)}_{i}, {\bf E}^{(k)}_{m}\right) / \tau\right)}
\end{align*}
}
\noindent where $\mathcal{M}(v_i)$ denotes a set negative samples, $\operatorname{cos}$ denotes the cosine similarity, and $\tau$ denotes the temperature parameter. In practice, following~\cite{chen2020simple,oord2018representation, khosla2020supervised}, $\mathcal{M}(v_i)$ is randomly sampled from $\mathcal{V}$. When the dataset is relatively small, the entire set $\mathcal{V}$ can serve as the set of negative samples. 

With the combined loss $\mathcal{L}$, the supervision signals are more robust to structural noises since graph information can be adaptively adjusted with $\alpha$. In the extreme case when the graph is totally unreliable, the balancing parameter $\alpha$ can be set to 0, making the supervised signals learned in $\mathcal{L}_{l}$ free of the effect of structural noise. The parameters ${\bf W}^{(k)}$ are learned by minimizing the overall decoupled loss $\mathcal{L}$, which generates representations and predictions applied in processes of \emph{node selection} and \emph{graph cleaning}.
\subsubsection{Node Selection}\label{sec:selection}
Here, \rrcl is expected to select $S$ nodes from $\mathcal{V}_{pool}$ that can represent $\mathcal{V}_{pool}$ in the best way. To achieve this, following FeatProp ~\cite{wu2019active} and other core-set approaches ~\cite{sener2017active, cai2017active, gao2018active, zhang2021alg}, we run the K-means with $S$ clusters utilizing the representations ${\bf E}^{(k)}$ produced by the \emph{representation learning} process. Then, we aim to select one node from each cluster. 
A straightforward way to do this is to select the most representative node from each cluster (the one close to the centroid). However, if the selected representative nodes are surrounded by noisy edges, inaccurate supervision signals will be propagated over the graph, which may even mislead the mode training. Hence, in addition to representativeness, we also care about the cleanliness of the neighborhood of candidate nodes. Based on this, we propose a novel cleanliness score that measures the risk of a node being influenced by noisy graph structures. When selecting nodes for labeling, we consider both the node representativeness score and the cleanliness score. 

As we have obtained a set of labeled nodes $\mathcal{V}^{(k-1)}$ from the previous iteration, we aim to avoid re-selecting them or any nodes that are adequately represented by this set. Thus, we first remove these nodes from the candidate pool. Next, we first describe the process of removing well-represented nodes. Then, we introduce the representativeness score and the robust cleanliness score. We conclude this section by describing how we use these metrics for node selection. \\
\noindent\textbf{Removing Well-Represented Nodes.} Clearly, nodes similar to ones in $\mathcal{V}_l^{(k-1)}$ should not be selected since they are already well-represented and will not provide further additional information. Therefore, before running the formal node selection process, we need to remove nodes that are close to $\mathcal{V}_l^{(k-1)}$ from $\mathcal{V}_{pool}$. In particular, we model the distance between a node $v_i \in \mathcal{V}_{pool}$ and the set $\mathcal{V}_l^{(k-1)}$ as follows. 
{\small
\begin{align*}
d(v_{i}, \mathcal{V}_l^{(k-1)})=\min_{v_j\in \mathcal{V}_l^{(k-1)}} d({v}_{i}, {v}_{j}),
\end{align*}
}
where $d({v}_{i}, {v}_{j})$ measures the Euclidean distance between node $v_i$ and $v_j$ using representations ${\bf E}^{(k)}$. We rank all nodes in $\mathcal{V}_{pool}$ in a non-decreasing order according to their distance to $\mathcal{V}^{k-1}_l$ and remove the top $|\mathcal{V}_l^{(k-1)}| \cdot h$ of them. $h>1$ is a hyper-parameter indicating how many nodes each labeled node covers. Note that nodes in $\mathcal{V}_l^{(k-1)}$ will always be removed as they have distance $0$.  After the removal, we denote the set of nodes left in the $\mathcal{V}_{pool}$ as the filtered candidate pool $\mathcal{V}_{filter}$. 

\noindent\textbf{Representativeness Score.} Since \rrcl selects one node per cluster, for each node in $\mathcal{V}_{filter}$, we define a representativeness score for each cluster. Let the centroid of the $s$-th cluster be denoted as $c_s$. For node $v_i\in \mathcal{V}_{filter}$, its representativeness score corresponding to the $s$-th cluster is defined as $r_{is} = 1/d(v_i, c_s)$,
where $d(v_i, c_s)$ measures the distance between $v_i$ and $c_s$. Intuitively, nodes with smaller distances are considered more representative.

\noindent\textbf{Cleanliness Score.} Nodes that are connected with clean edges often share similar features. Hence, we define the cleanliness score for a node $v_i\in \mathcal{V}_{filter}$ based on the feature similarity to its neighbors as $ cl_{i} = \sum_{v_j \in \mathcal{N}(v_i)} cos({\bf x}_n, {\bf x}_j)$,

where $\mathcal{N}(v_i)$ denotes the set of neighbors of node $v_i$ and $cos({\bf x}_n, {\bf x}_j)$ measures the cosine similarity between their original features.

\noindent\textbf{Node selection with Representativeness and Cleanliness Scores.} Next, \rrcl leverages the representativeness score and the cleanliness score together for node selection. Clearly, the representativeness score and the cleanliness score are at different scales and we care more about the relative relations between the candidate nodes in $\mathcal{V}_{filter}$. Hence, to combine these two scores, we rank the scores of candidate nodes and convert the scores into percentiles following \cite{zhang2017active, cai2017active}. Specifically, for the $i$-th cluster, we rank all nodes in $\mathcal{V}_{filter}$ according to their representativeness score corresponding to the $s$-th cluster in a non-increasing order and obtain the percentile for each node. We denote the percentile for node $v_i\in \mathcal{V}_{filter}$ corresponding to $i$-th cluster as $\hat{r}_{is}$. Similarly, we obtain the percentile based on the cleanliness score, which is denoted as $\hat{cl}_i$ for node $v_i$. With these two percentiles, we select nodes for labeling as follows. 
{\small
 \begin{align*} 
 \mathcal{V}_{select}^{(k)} = \bigcup_{s =1}^{S} \{\argmin_{v_i \in \mathcal{V}_{filter}} \space \hat{r}_{is} + \beta \cdot  \hat{cl}_{i} \},
\end{align*}
}
where $\beta$ balances these two kinds of information, and we select the node with the largest combined score from each cluster. The labels of selected nodes $ \mathcal{V}_{select}^{(k)}$ are then queried from the oracle $\mathcal{O}$. Finally, we include the newly selected node set $ \mathcal{V}_{select}^{(k)}$ to the previous labeled set $\mathcal{V}_l^{(k-1)}$ expand the labeled set as $\mathcal{V}_l^{(k)} = \mathcal{V}_{select}^{(k)} \bigcup \mathcal{V}_l^{(k - 1)}$.

\subsubsection{Graph Cleaning}\label{sec:gclean}
In this part, \rrcl is designed to clean the graph structure by identifying and down-weighting the noisy edges in the graph with an edge-predictor. However, building such an edge-predictor in the AL setting is extremely challenging as labels are scarce. Specifically, it requires querying two nodes from the oracle to verify whether an edge is noisy or clean. In light of this challenge, we propose to construct a training set with pseudo labels of edges utilizing the representations ${\bf E}^{(k)}$. We then utilize this training set to train an edge-predictor, which is utilized for cleaning the graph. Next, we first describe the training set construction, and then introduce details on utilizing the edge-predictor for graph learning.

\noindent{\bf Edge Training Set Construction.} \label{sec:edgeclean}
To produce pseudo labels for edges, we first produce probability logits for all nodes utilizing $MLP_2$ and ${\bf E}^{(k)}$ obtained from the \emph{representation learning} component. We denote the vector of logits for node $v_i\in \mathcal{V}$ as ${\bf p}^{(k)}_i$. Note that for labeled nodes, we use their one-hot label vectors to replace the logits. We first obtain the pseudo labels for all nodes from the logits. Specifically, for each node $v_i$, the index corresponding to the largest dimension in the logits  ${\bf p}^{(k)}_i$ is treated as the pseudo label, denoted as $\hat{y}_i$. Intuitively, we consider an edge to be ``clean'' when its two nodes share the same label with high confidence. Therefore, the nodes in the following set are treated as positive samples. 
{\small
\begin{align*}
& \mathcal{E}^{pseudo}_{+} = \{   e_{ij} \in \mathcal{E} \mid  \space v_i\in \mathcal{V}_{con}, v_j \in \mathcal{V}_{con}, \hat{y}_i= \hat{y}_{j}\}
\end{align*}
}
where $\mathcal{V}_{con} = \{v_i \in \mathcal{V} \mid  {\bf p}_{i}[\hat{y}_i] \geq \kappa \}$ denote the set of nodes with confident pseudo labels, and $\kappa$ denotes a threshold of confidence. On the other hand, we consider an edge as ``noisy'' when its two nodes have different pseudo labels, as formulated below. 
{\small
\begin{align*}
& \mathcal{E}^{pseudo}_{-} = \{   e_{ij} \in \mathcal{E} \mid  \space v_i\in \mathcal{V}, v_j \in \mathcal{V}, \hat{y}_i \not= \hat{y}_{j}\}.
\end{align*}
}
Note that, we do not enforce the confidence constraint to the negative samples, since the edge is still highly likely to be negative even if the confidence of node pseudo labels is extremely high (larger than $\kappa$). We tried to enforce the constraint, which turned out to affect the overall performance insignificantly.  

\noindent\textbf{Edge-predictor.} With the training data defined previously, we train an edge-predictor. The probability of an edge being clean is modeled as follows.
{\small
\begin{align*}
    p(e_{ij} = 1) = \sigma({\bf z}_i^{\top} {\bf z}_j) \text{ with } {\bf z}_i = MLP_3({\bf x}_i; {\bf W}^{(k)}_3),
\end{align*}
}
where $MLP_3$ maps the original features into a representation space that is suitable for the probability estimation. The edge predictor is trained by maximizing the following probability.
{\small
\begin{align*}
    P_{edge} = \prod\limits_{e_{ij}\in \mathcal{E}^{pseudo}_{+}} p(e_{ij} = 1) \cdot \prod\limits_{e_{ij}\in \mathcal{E}^{pseudo}_{-}}  (1-p(e_{ij} = 1)).
\end{align*}
}
In practice, instead of maximizing $P_{edge}$, we minimize the negative of its logarithm. Once we obtain $ {\bf W}^{(k)}_3)$, we infer the probability $p(e_{ij} = 1)$ for all edges $e_{ij}\in \mathcal{E}$ and update the edge weights in the adjacency matrix as follows.
{\small
\begin{align} 
{\bf A}_{i j}^{(k)}= \begin{cases} p(e_{ij} = 1), & {e}_{ij} \in \mathcal{E} \\ 0, & \text { others. }\end{cases}\label{eq:prob_graph}
\end{align} 
}
The probabilistic adjacency matrix ${\bf A}_{i j}^{(k)}$ defines a distribution for the discrete clean graph $\mathcal{G}$. In our case, we adopt the weighted graph defined by ${\bf A}_{i j}^{(k)}$ as $\mathcal{G}^{(k)}$ for the \emph{representation learning} in the following iteration. Note that $\mathcal{G}^{(k)}$ is the expectation of the distribution defined by ${\bf A}_{i j}^{(k)}$.

\section{\rrcl as an EM algorithm}  \label{sec:gcl_em}
In this section, we show that the \rrcl framework can be understood from the perspective of an expectation-maximization (EM) algorithm. We first briefly introduce the concepts of the EM algorithm. Then, we explain how our framework can be formulated as an instance of the EM algorithm.

\subsection{EM algorithm} 
An EM \cite{dempster1977maximum} is an iterative method used in machine learning for parameter estimation in probabilistic models that involve latent variables. The EM operates on a joint distribution $p(\mathbf{U}, \mathbf{z} \mid \boldsymbol{\theta})$ over the observed variable $\mathbf{U}$, the unobserved latent variables $\mathbf{z} \in \mathcal{Z} $ and model parameters $\boldsymbol{\theta}$. The goal is to maximize the likelihood function $p(\mathbf{U} \mid \boldsymbol{\theta})$ with respect to $\boldsymbol{\theta}$. Given $N$
 observations $\left\{\mathbf{u}_i\right\}_{i=1}^N \text { of } \boldsymbol{X}$, The EM typically follows a two-step process: \\
\noindent \textbf{E-step}: The E-step is to compute the expected value of the likelihood function given the observed data, the posterior distribution of the latent variables, and updated parameters $\boldsymbol{\theta}^{\text{old}}$
{\small
\vspace{-0.1cm}
\begin{align*} 
\mathcal{Q}\left(\boldsymbol{\theta}; \boldsymbol{\theta}_{old}\right) = \frac{1}{N} \sum_{i=1}^N \int_{\mathcal{Z}} p\left(\mathbf{z} \mid \mathbf{u}_i\right) \cdot \log p\left(\mathbf{u}_i, \mathbf{z}\right) \mathrm{d} \mathbf{z}.
\end{align*} 
}\noindent \textbf{M-step}: The M-step is to update the parameters by maximizing the function $\mathcal{Q}\left(\boldsymbol{\theta};\boldsymbol{\theta}_{old}\right)$. Typically, the maximization in the M-step is conducted through gradient-based methods. In large-scale settings, computing the full gradient can be computationally costly. Stochastic Expectation Maximization~\cite{zhu2017high, balakrishnan2017statistical, papandreou2015weakly} addresses this challenge by employing stochastic gradient methods, where the gradient is estimated using a subset of the data. 

\begin{figure*}[!htp]%
\vspace{-1.5cm}
\captionsetup[subfloat]{captionskip=0mm}
     \centering
     \subfloat[\cora]{{\includegraphics[width=0.28\linewidth]{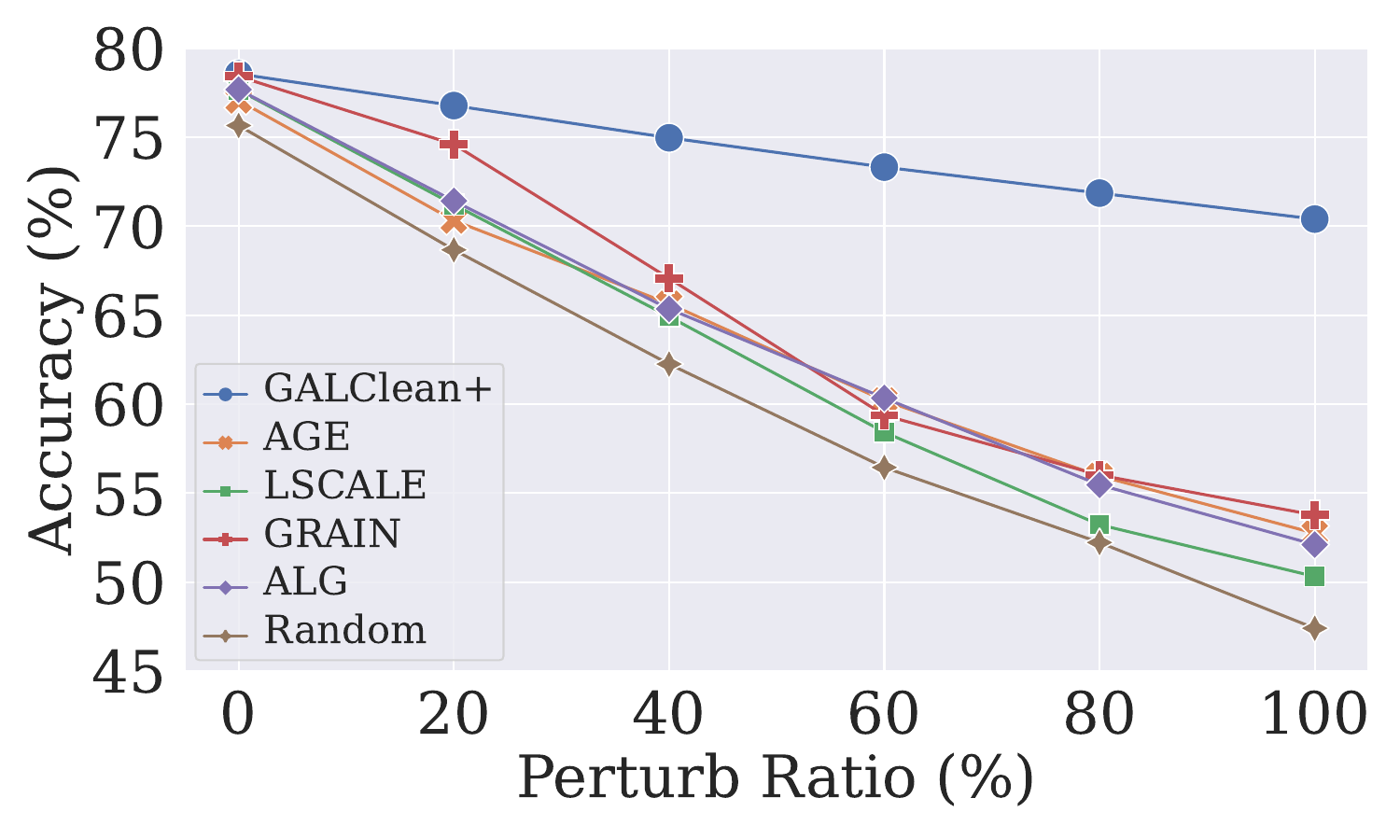}}}%
     \subfloat[\citeseer]{{\includegraphics[width=0.28\linewidth]{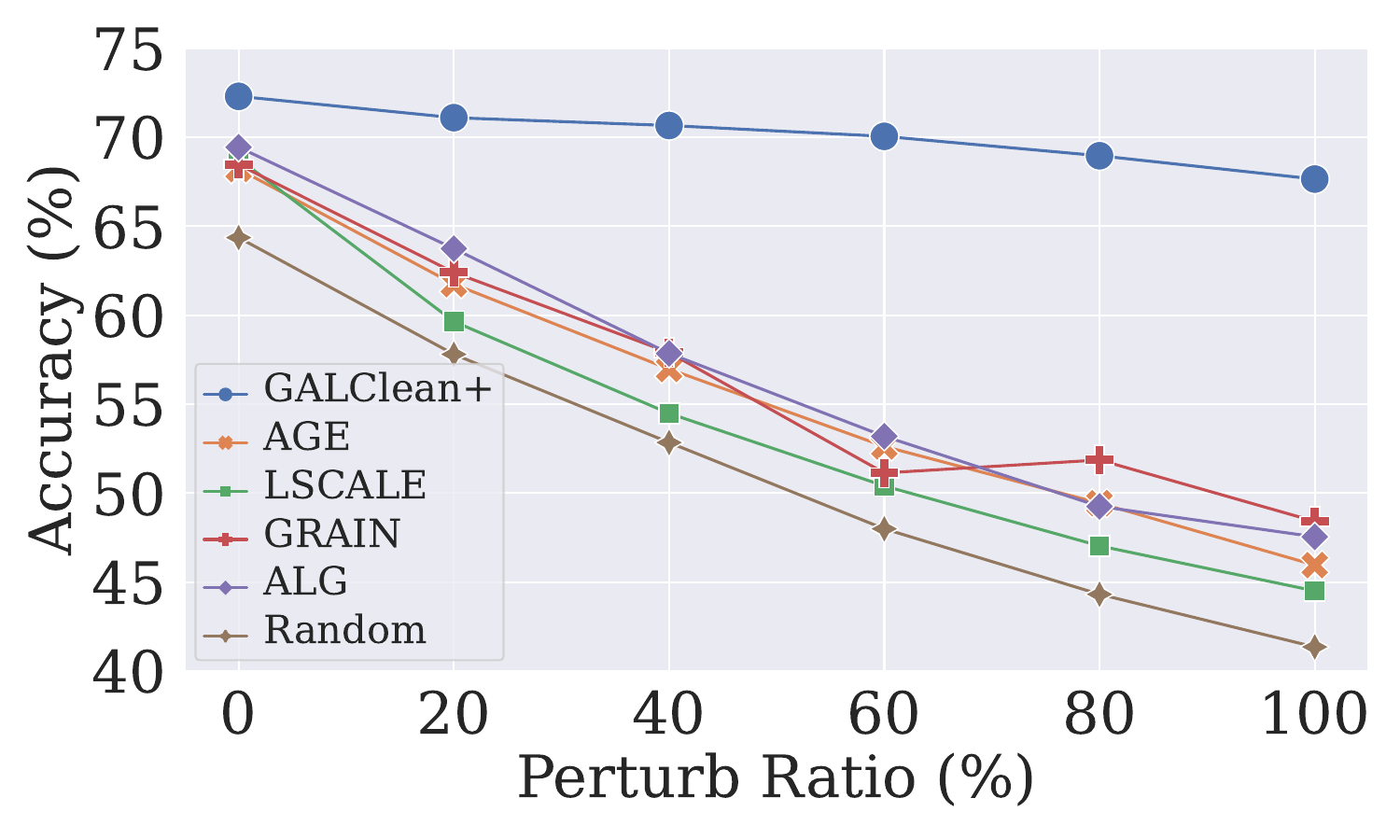} }}%
    \subfloat[\pubmed]{{\includegraphics[width=0.28\linewidth]{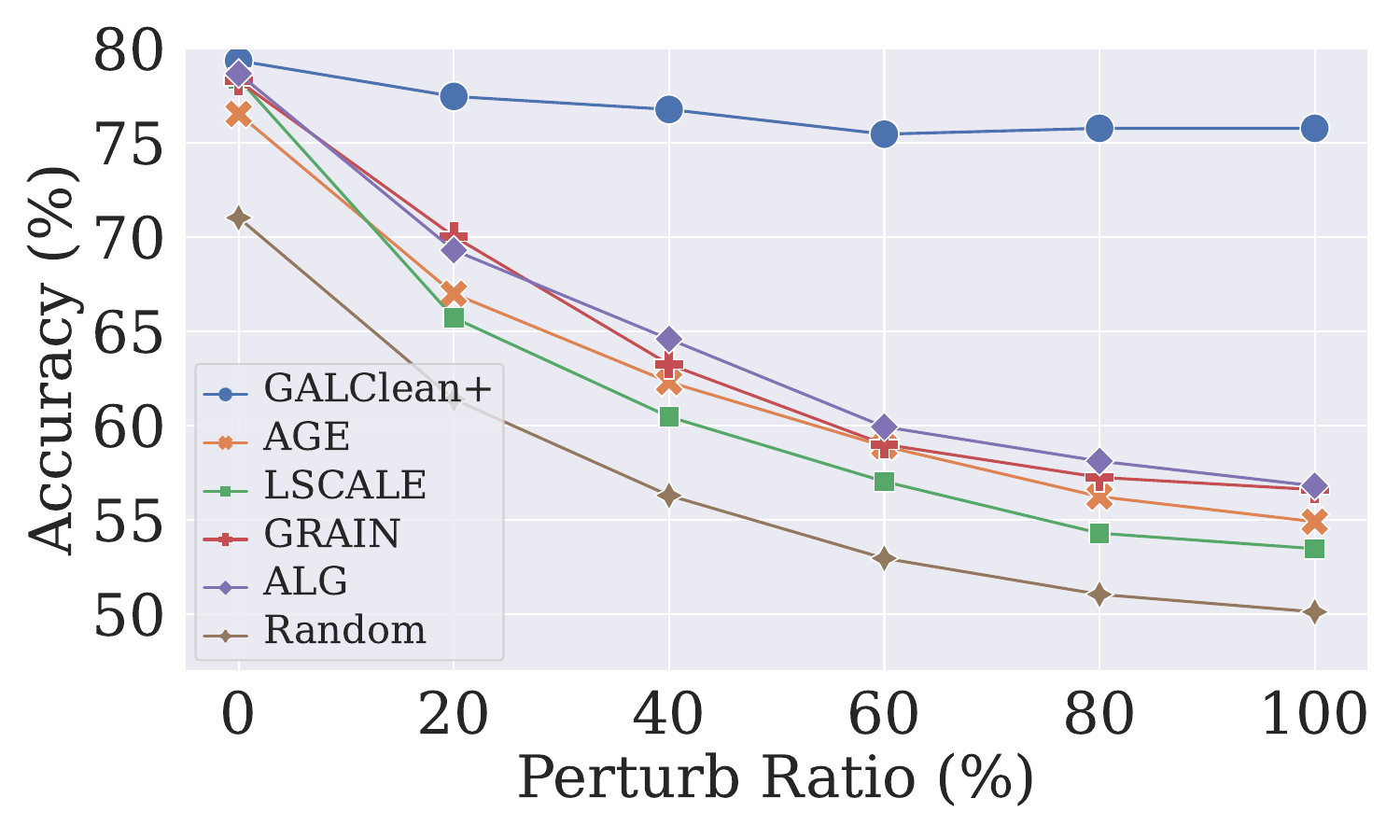} }}%
        ~\\%
    \subfloat[\photo]{{\includegraphics[width=0.28\linewidth]{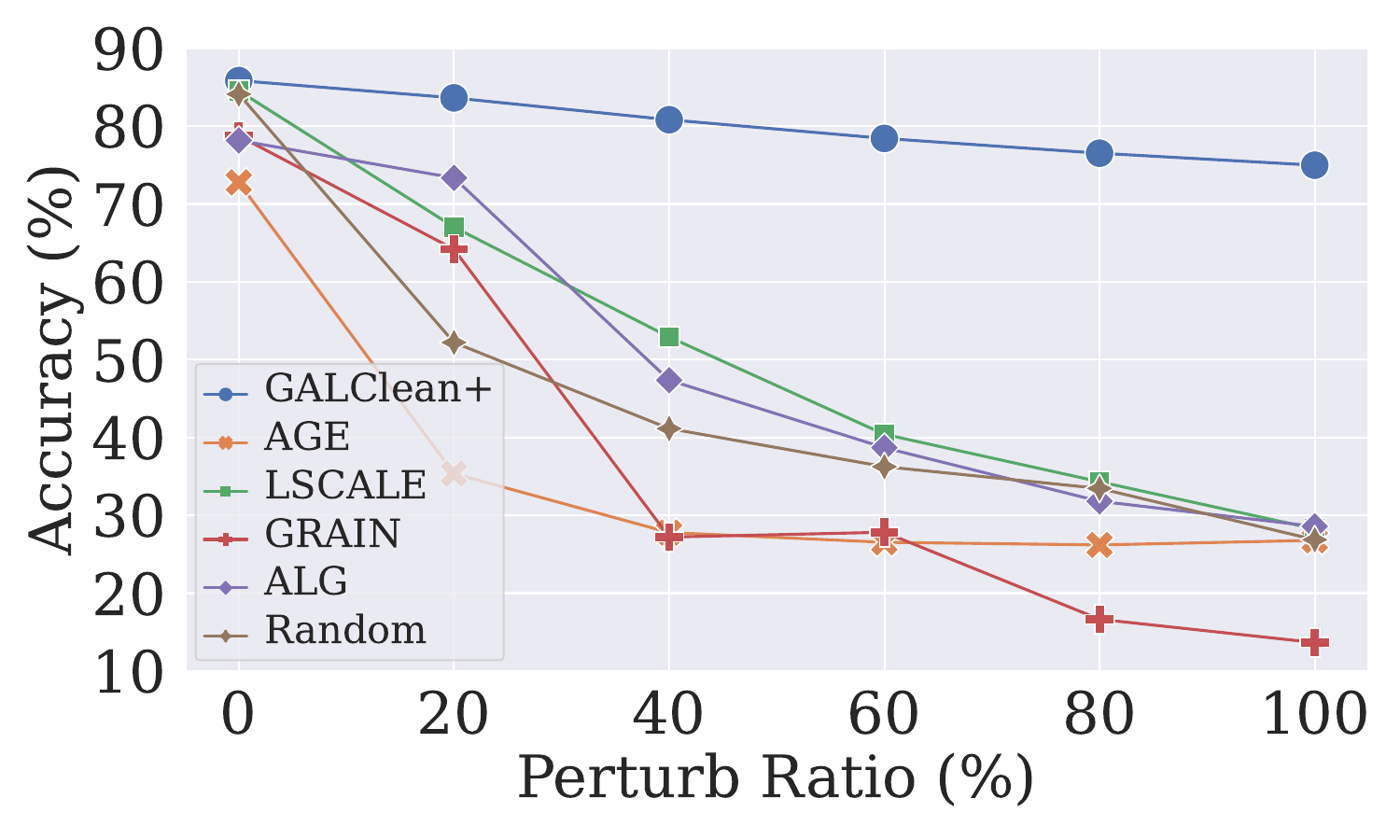} }}%
    \subfloat[\comp]{{\includegraphics[width=0.28\linewidth]{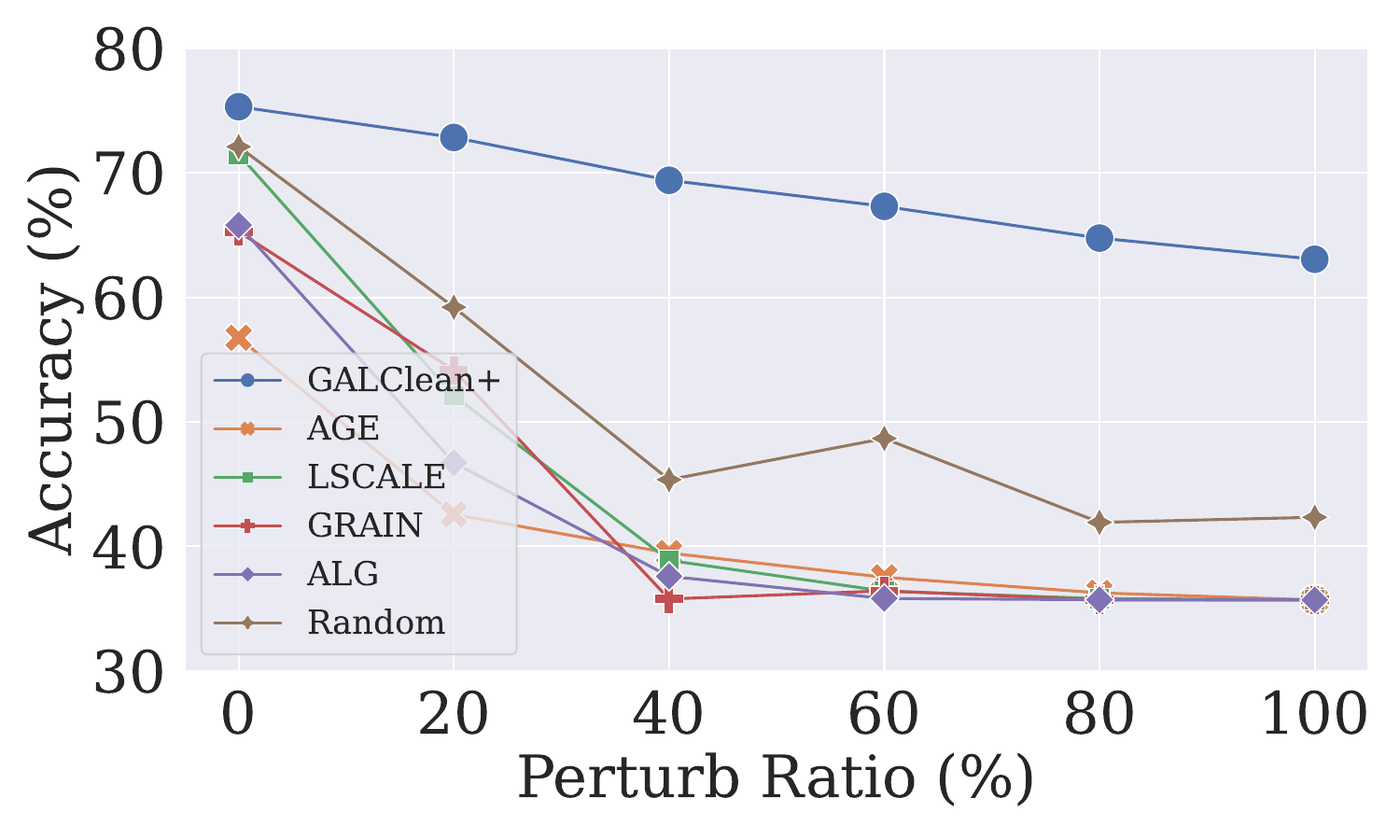} }}%
    \subfloat[\cocs]{{\includegraphics[width=0.28\linewidth]{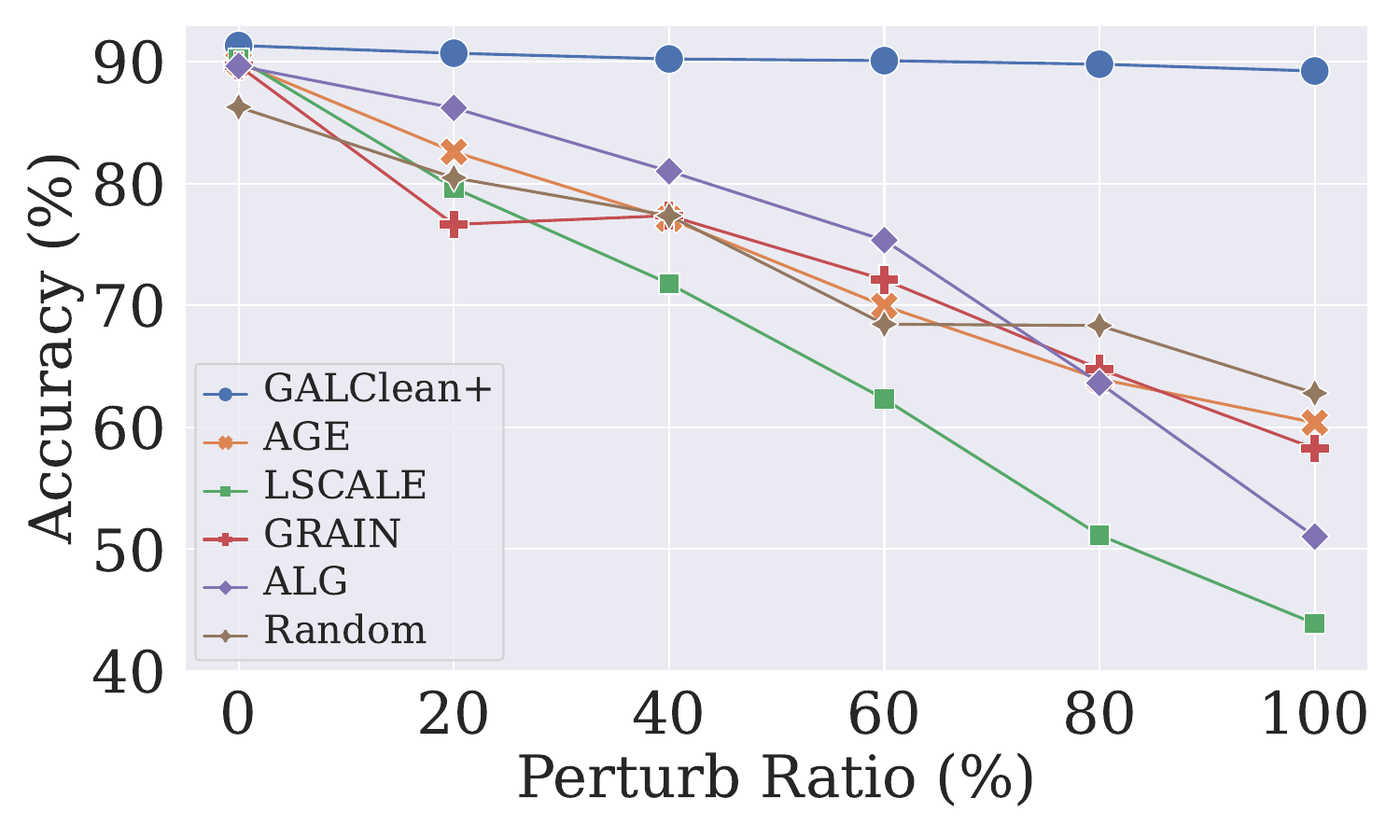} }}%
    \caption{\small{Active Learning Performance under Random Edge-Adding Attacks}}
    \label{fig:random_result}
    \vspace{-0.5cm}
\end{figure*}

\subsubsection{Interpret \rrcl as a Stochastic EM }
In our setting, the original graph structure is provided but with noisy edges. Since the clean graph structure is unknown, we treat it as the latent variable. Ideally, if we were able to observe labels for all nodes, they could serve as the observed variables in a standard EM. In this case, the goal of EM is to maximize the following marginal log-likelihood of all nodes in $\mathcal{V}$ with their corresponding labels observed:
{\small
\begin{align*} 
 \sum_{v_i \in \mathcal{V} } \ln p( {\bf y}_i , \mathbf{X} \mid \boldsymbol{\theta}) = \sum_{v_i \in \mathcal{V} } \int \ln p( {\bf y}_i, \mathbf{X}, \mathbb{G} \mid \boldsymbol{\theta}) d\mathbb{G}.
\end{align*} 
}

where $p( {\bf y}_i, \mathbf{X}, \mathbb{G}, \mid \boldsymbol{\theta})$ is the likelihood for node $v_i$ with label ${\bf y}_i$ given the latent graph $\mathbb{G}$, and $\boldsymbol{\theta}$ corresponds to the model parameters. However, in our case, only a very small subset of nodes are observed with labels. Hence, we adopt stochastic gradient methods to optimize the log-likelihood in the M-step. In particular, the iterative process of \rrcl can be regarded as an instance of a stochastic EM. We utilize the $k$-th iteration of the \rrcl to illustrate the corresponding E and M-steps. \\
\noindent \textbf{E-step}: In the E-step, we aim to obtain the expectation of likelihood function given the observed data, the updated distribution of latent variables, and the updated parameters:
{\small
\begin{align}
&\mathcal{Q}\left(\boldsymbol{\theta}; \boldsymbol{\theta}^{(k-1)}\right) = \nonumber \\
& \sum_{v_i \in \mathcal{V}} \int p\left(\mathbb{G} \mid \mathbf{X}, {\bf y}_i, \boldsymbol{\theta}^{(k-1)}\right) \ln p({\bf y}_i, \mathbf{X},  \mathbb{G} \mid \boldsymbol{\theta}) d\mathbb{G}, \label{eq:expected-log-like}
\end{align}
}where $ \boldsymbol{\theta}^{(k-1)}$ refers to the model parameters estimated at the ($k-1$)-th iteration. $p\left(\mathbb{G} \mid \mathbf{X}, {\bf y}_i, \boldsymbol{\theta}^{(k-1)}\right)$ is the posterior distribution of the latent graph $\mathbb{G}$, which is described by the probabilistic adjacency matrix in Eq.~\eqref{eq:prob_graph}. Computing the expectation in  Eq.~\eqref{eq:expected-log-like} is prohibitively expensive. Therefore, we approximate the entire posterior distribution using a Dirac delta function ($\delta$ distribution), a method known as variational approximation. The optimal $\delta$ distribution is defined at the Maximum a posteriori (MAP) of $\mathbb{G}$~\cite{beal2003variational}. In particular, in our case, the mass of $\delta$ distribution is concentrated at the expectation of the posterior distribution $\mathbb{G}^{(k-1)}$ (obtained in \emph{graph cleaning}) and has 0 mass anywhere else. With the $\delta$ distribution, we approximate Eq.~\eqref{eq:expected-log-like} as follows.
{\small
\begin{equation}
\mathcal{Q}\left(\boldsymbol{\theta}; \boldsymbol{\theta}^{(k-1)}\right) = \sum_{v_i \in \mathcal{V}} \log p({\bf y}_i , \mathbf{X},  \mathbb{G}^{(k-1)} \mid \boldsymbol{\theta}) \label{eq: approx-expec}
\end{equation}
}
In conclusion, the E-step corresponds to the \emph{graph cleaning} component in \rrcl.

\noindent \textbf{M-step}: In the M-step, we maximize Eq.~\eqref{eq: approx-expec}. Due to the limited labeled data, we optimize it and obtain the updated parameters $\boldsymbol{\theta}^{(k)}$ with stochastic gradient estimated on $\mathcal{V}_l^{(k-1)}$. The M-step corresponds to the \emph{representation learning} component in the \rrcl framework. Specifically, $\boldsymbol{\theta}^{(k)}$ summarizes all parameters in Section~\ref{sec:gclean} including ${\bf W}_1^{(k)}$ and ${\bf W}_2^{(k)}$. Note that the \emph{data selection} also plays an important role in the M-step, as it gradually provides better $\mathcal{V}_l^{(k-1)}$ for a more accurate estimation of the full gradient.

Next, we explain how the \emph{graph cleaning} and \emph{data selection} enhance each other from the perspective of the stochastic EM algorithm: (a) The EM guarantees that the log-likelihood value increases with each iteration until convergence, leading to a progressively improved fit of the representation model. This improvement ensures that both the AL and graph cleaning processes are fully leveraged based on the most recent and reliable information obtained thus far, which leads to high-quality \emph{data selection}; and (b)  Moreover, since only a batch of observed data is used to optimize the likelihood function during the M-step, the gradient derived from the batch data may exhibit high variance, especially when the size of observations is small. The proposed \emph{data selection} strategy focuses on selecting data with representativeness and cleanliness, which helps to obtain a more reliable mini-batch gradient that is better aligned with the one derived from fully labeled nodes with the clean graph. This alignment ensures that the Stochastic EM is updated in a more unbiased manner, thereby enhancing the overall effectiveness and accuracy of the model, which, in turn, helps the inference of the latent graph in the \emph{graph cleaning} component (E-step). 

\subsubsection{\galp} \label{sec:refinement}
As described in the previous section, \rrcl can be understood as an instance of a stochastic EM. A natural idea to extend the \rrcl is to run a few more iterations of EM even after the labeling budget is exhausted, which may help further clean the graph. Hence, we propose an enhanced version of \rrcl named \galp. In particular, after $K$ iterations of \rrcl, we run out of the labeling budget and obtain $\mathcal{V}_l^{(K)}$. We continue the EM for a few more iterations, where, in the $M$-step, we always use $\mathcal{V}_l^{(K)}$ to calculate the gradient. We also treat the unlabeled data $\mathcal{V}/ \mathcal{V}_l^{(K)}$ as unobserved latent variables and involve them in the remaining EM iterations. 

In the end, we present a detailed \underline{analysis of time} \underline{complexity} for \rrcl and \galp in Section 2 of the \textbf{supplementary file}.

\section{Experiments}\label{sec:exper}
In this section, we conduct experiments to assess the effectiveness and robustness of our framework in two noisy scenarios or attacking cases. After that, we test and highlight the significance of its key design modules by conducting ablation studies and parameter analyses. Notably, we include the details of the \underline{datasets}, \underline{baselines}, \underline{graph noise/attack} \underline{generation mechanisms}, and \underline{experimental settings} in the \textbf{supplementary materials}.

\begin{figure*}[!htp]%
\vspace{-1.5cm}
\captionsetup[subfloat]{captionskip=0mm}
     \centering
     \subfloat[\cora]{{\includegraphics[width=0.27\linewidth]{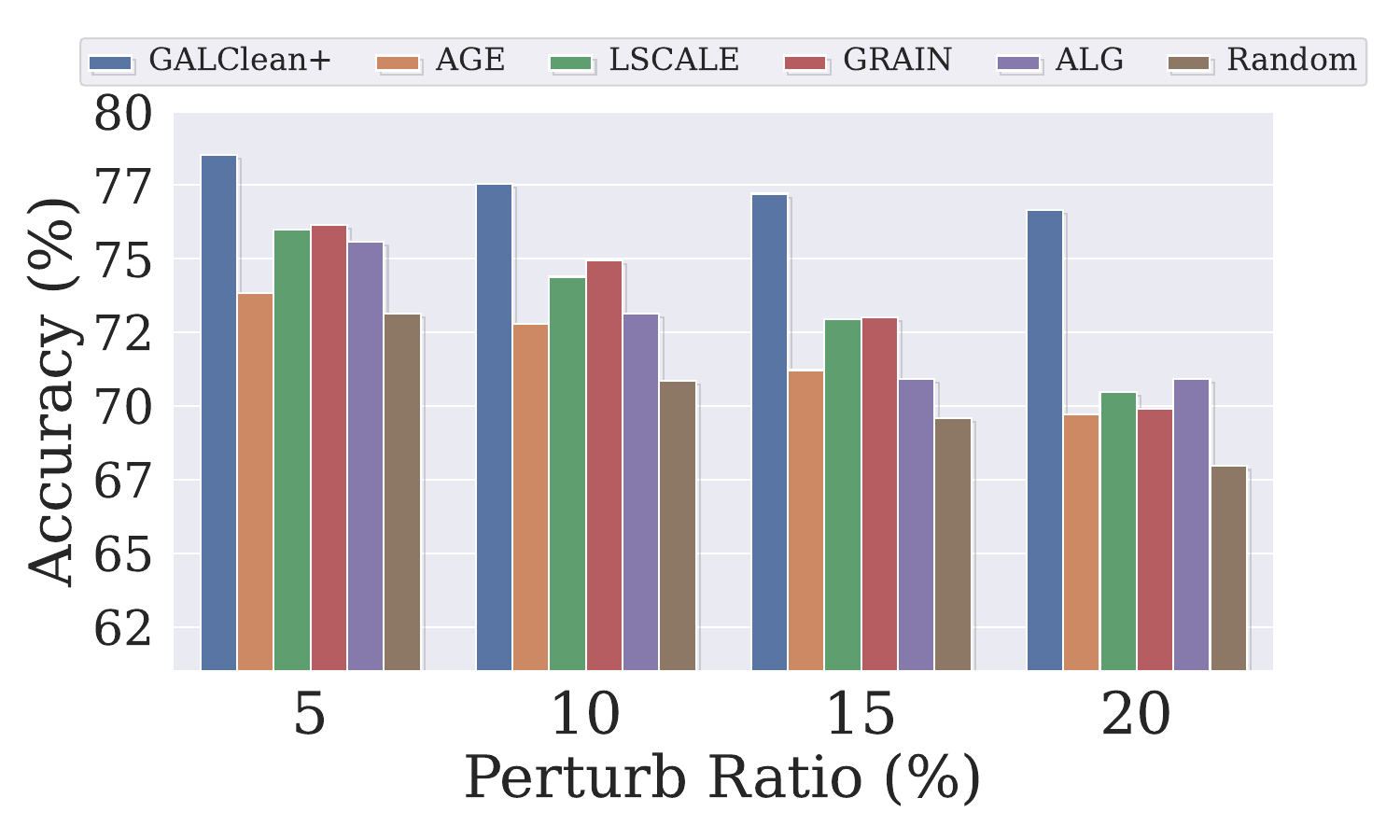}}}%
     \subfloat[\citeseer]{{\includegraphics[width=0.27\linewidth]{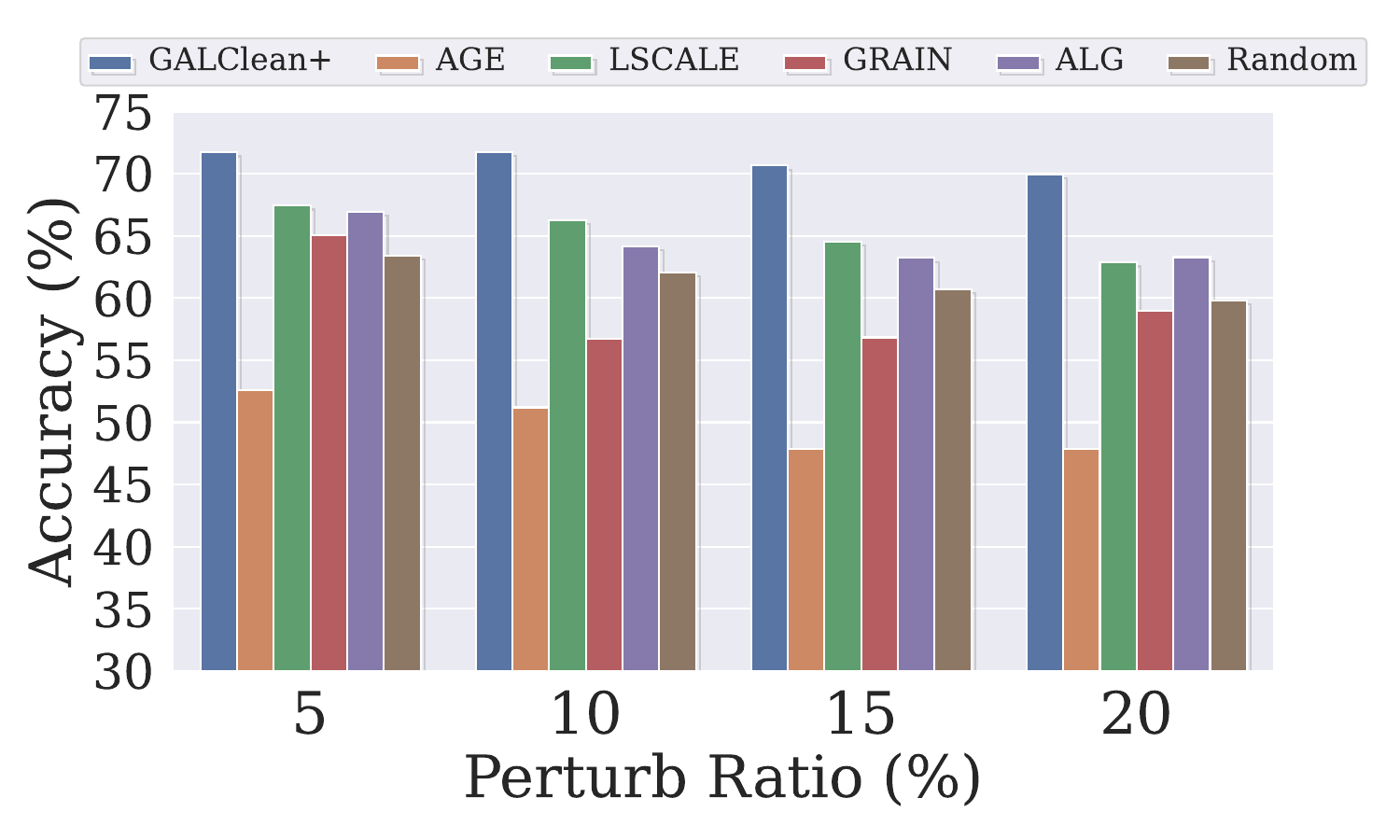} }}%
    \subfloat[\pubmed]{{\includegraphics[width=0.27\linewidth]{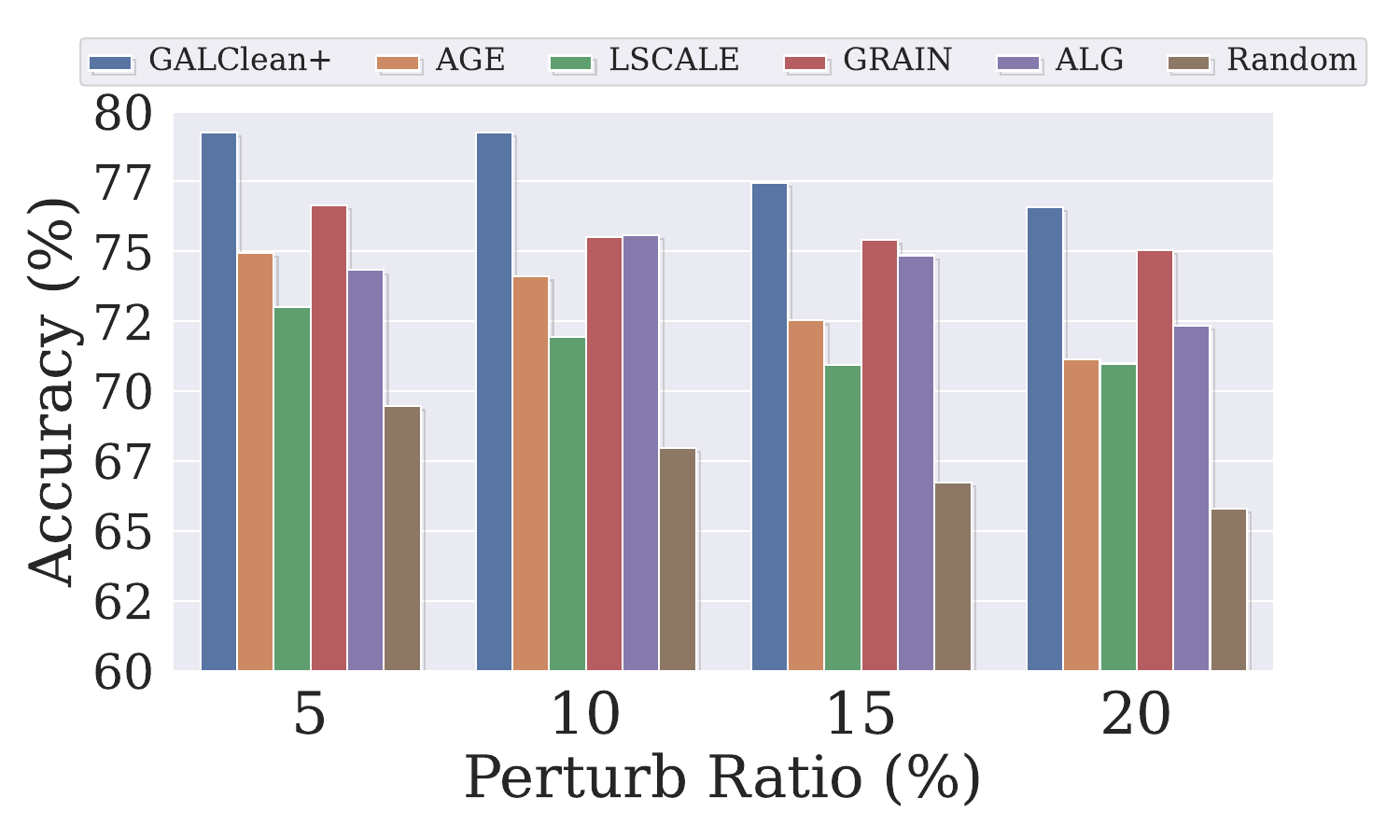} }}%
    ~\\%
    \subfloat[\photo]{{\includegraphics[width=0.27\linewidth]{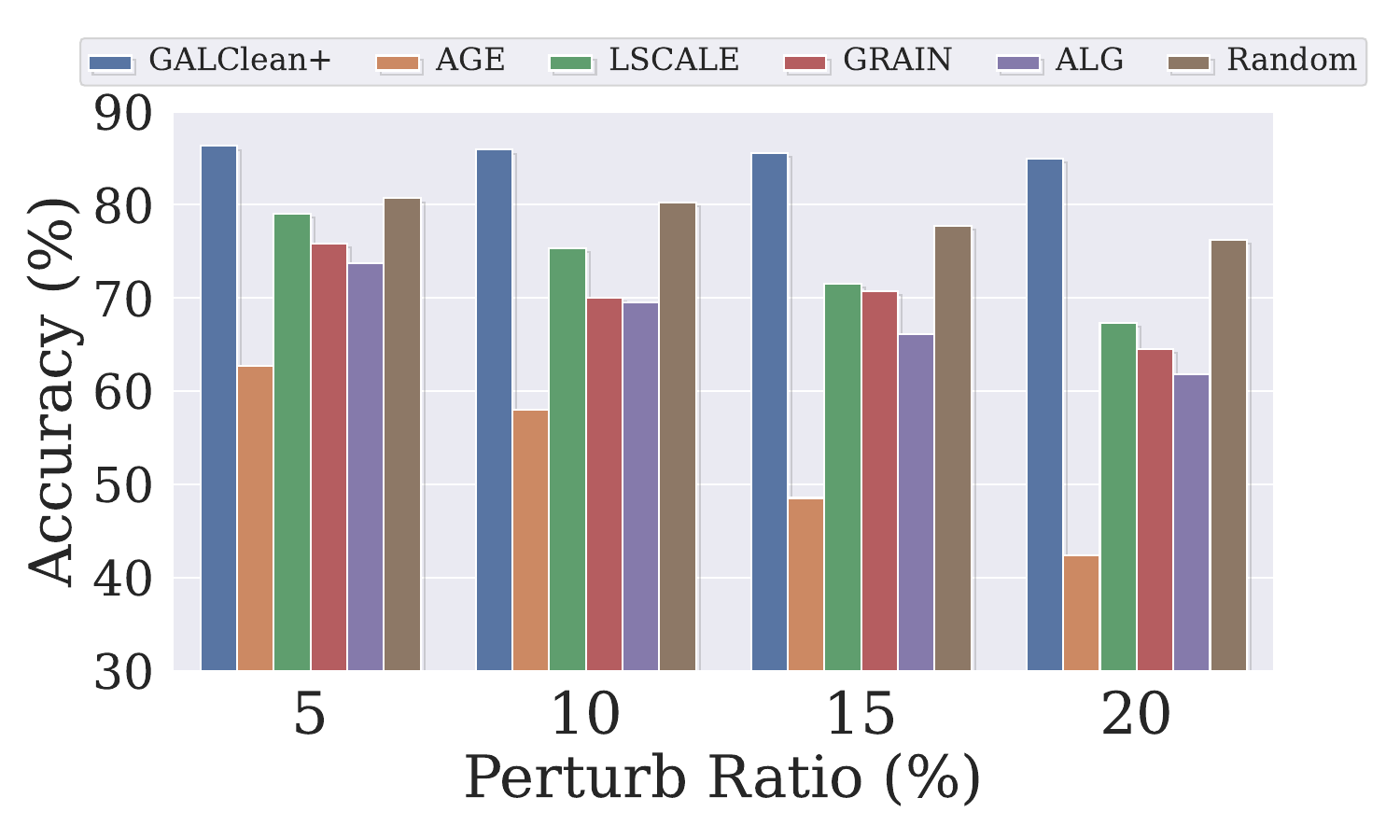} }}%
    \subfloat[\comp]{{\includegraphics[width=0.27\linewidth]{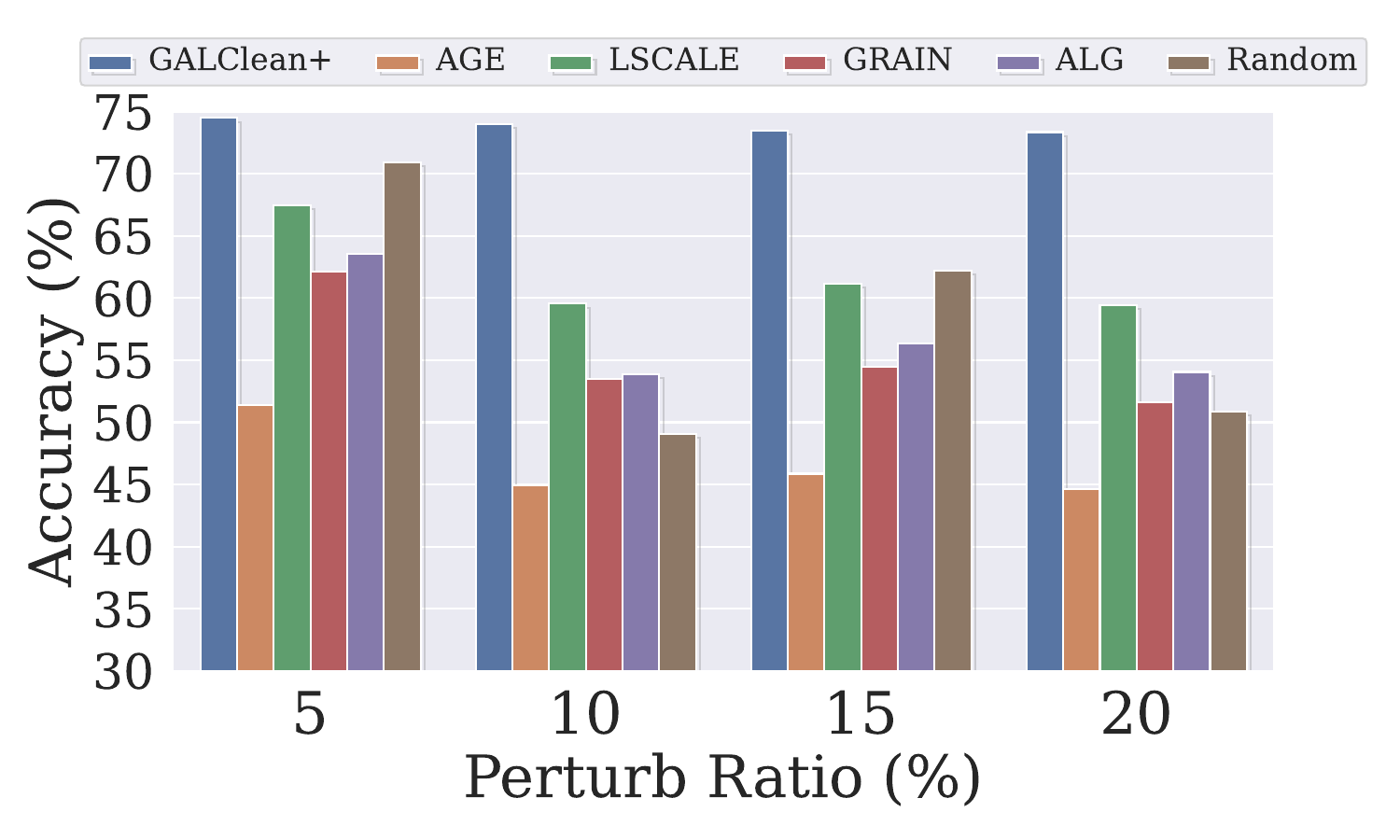} }}%
    \subfloat[\cocs]{{\includegraphics[width=0.27\linewidth]{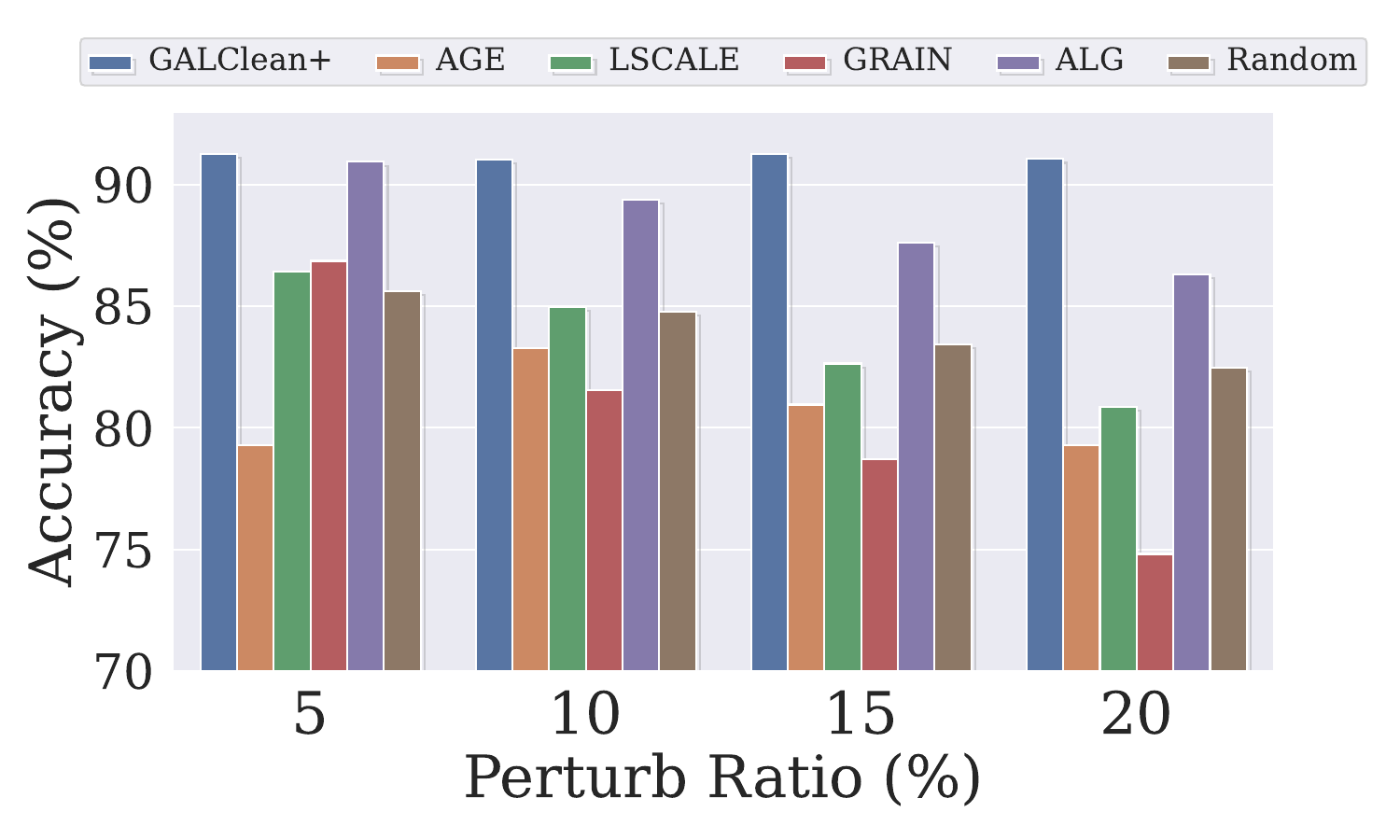} }}%
    \caption{\small{Active Learning Performance under Unsupervised Adversarial Attacks}}\label{fig:node_classification_performance}
    \label{fig:unsupervised}
 \vspace{-0.5cm}
\end{figure*}

\subsection{Main Results}
Here, we assess \galp's performance under various types and levels of attacks. In particular, we adopt \emph{Random Edge-Adding Attack} and \emph{Unsupervised Adversarial Attack} to perturb graphs at different levels. The perturbation process of these attacks can be found in Section 1 of supplementary file. Next, We present the major results of \galp against comprehensive baselines with a thorough analysis. 

\noindent \textbf{\emph {Random Edge-Adding Attacked Graphs.}} The results on graphs with $n\%$ random edge-adding noises are shown in Figure~\ref{fig:random_result}. It can be seen that \galp outperforms various baselines by a large margin on all six datasets, which indicates
the effectiveness of \galp on selecting high-quality nodes while cleaning the graph. 

\noindent \textbf{\emph{Unsupervised Adversarial Attacked Graphs.}} In this part, we investigate how robust \galp is when graphs are perturbed by $n\%$ unsupervised adversarial attacks. The testing accuracy of \galp and baselines are reported in Figure~\ref{fig:unsupervised}. Once again, \galp outperformed all baselines, which demonstrates the superiority of \galp even under a sophisticated unsupervised attack. 

\subsection{Ablation Study and Parameter Analysis}
In this subsection, we scrutinize the design modules of \galp through ablation study and parameter analysis. We report the results under $100\%$ random edge-adding noise setting on citation datasets as the observations on other settings and datasets are similar. Additionally, we investigated how \galp performs with limited labelling budgets. This \underline{budget sensitivity analysis} is attached in Section 3 of the \textbf{supplementary file}.

\subsubsection{Effectiveness of the Cleanliness-based Node Selection Strategy}
For better understanding how the proposed selection strategy, particularly the necessity of the node cleanliness metric introduced in Section~\ref{sec:selection}, we run \galp
with and without the cleanliness score (C-score). 
 
All other parameters and settings were kept the same. The results are shown in Table~\ref{tab:cleaniess}. As we can observe from the table, on all three datasets, the cleanliness score helped improve the performance significantly, which shows the effectiveness of the proposed cleanliness-based data selection strategy for active learning tasks on noisy graphs. 
\vspace{-0.3cm}
\begin{table}[H]
\caption{\small{Impact of the Use of Cleanliness Score on Testing Accuracy (\%) (${\color{red} \uparrow\uparrow}$: increase $>$ 1\%)}}
\begin{adjustbox}{width=0.5\textwidth}
\begin{tabular}{ccccc}
\toprule
\textbf{Setting}   & \cora & \citeseer & \pubmed \\
\midrule
w/ C-Score  &  \textbf{70.40$\pm$1.35 ${\color{red} \uparrow\uparrow}$}  & \textbf{67.65$\pm$1.52 ${\color{red} \uparrow\uparrow}$}   & \textbf{75.76$\pm$2.32 ${\color{red} \uparrow\uparrow}$} \\
w/o C-Score   &  69.21$\pm$1.47 & 64.37$\pm$2.68  & 73.31$\pm$3.91  \\
\bottomrule
\end{tabular}
\end{adjustbox}
\label{tab:cleaniess}
\vspace{-0.5cm}
\end{table}

\subsubsection{\galp versus. \rrcl} To clearly illustrate the improvement from the additional EM iterations in \galp (introduced in Section~\ref{sec:refinement}), we compare the performance of \rrcl and  \galp side by side, which is summarized in Table~\ref{tab:refinement}. The results show that \galp consistently outperforms \rrcl, which confirms the use of extra EM iterations in purifying graph structures and enhancing data selection quality.

\begin{table}[H]
\caption{Tesing Accuracy (\%) of \galp and \rrcl (${\color{red} \uparrow}$: increase $>$ 0.5\%; ${\color{red} \uparrow\uparrow}$: increase $>$ 1\%)} 
\begin{adjustbox}{width=0.5\textwidth}
\begin{tabular}{ccccc}
\toprule
\textbf{Model}   & \cora & \citeseer & \pubmed \\
\midrule
\galp  &  \textbf{70.40$\pm$1.35 ${\color{red} \uparrow}$}  & \textbf{67.65$\pm$1.52${\color{red} \uparrow\uparrow}$}   & \textbf{75.76$\pm$2.32 ${\color{red} \uparrow}$} \\
\rrcl  &  69.81$\pm$1.14 & 65.99$\pm$1.62 & 75.08$\pm$2.29 \\
\bottomrule
\end{tabular}
\end{adjustbox}
\label{tab:refinement}
\vspace{-0.5cm}
\end{table}

\subsubsection{Effectiveness of Thresholding with  $\kappa$}
\begin{figure}[!htp]
     \centering
     {{\includegraphics[width=0.7\columnwidth]{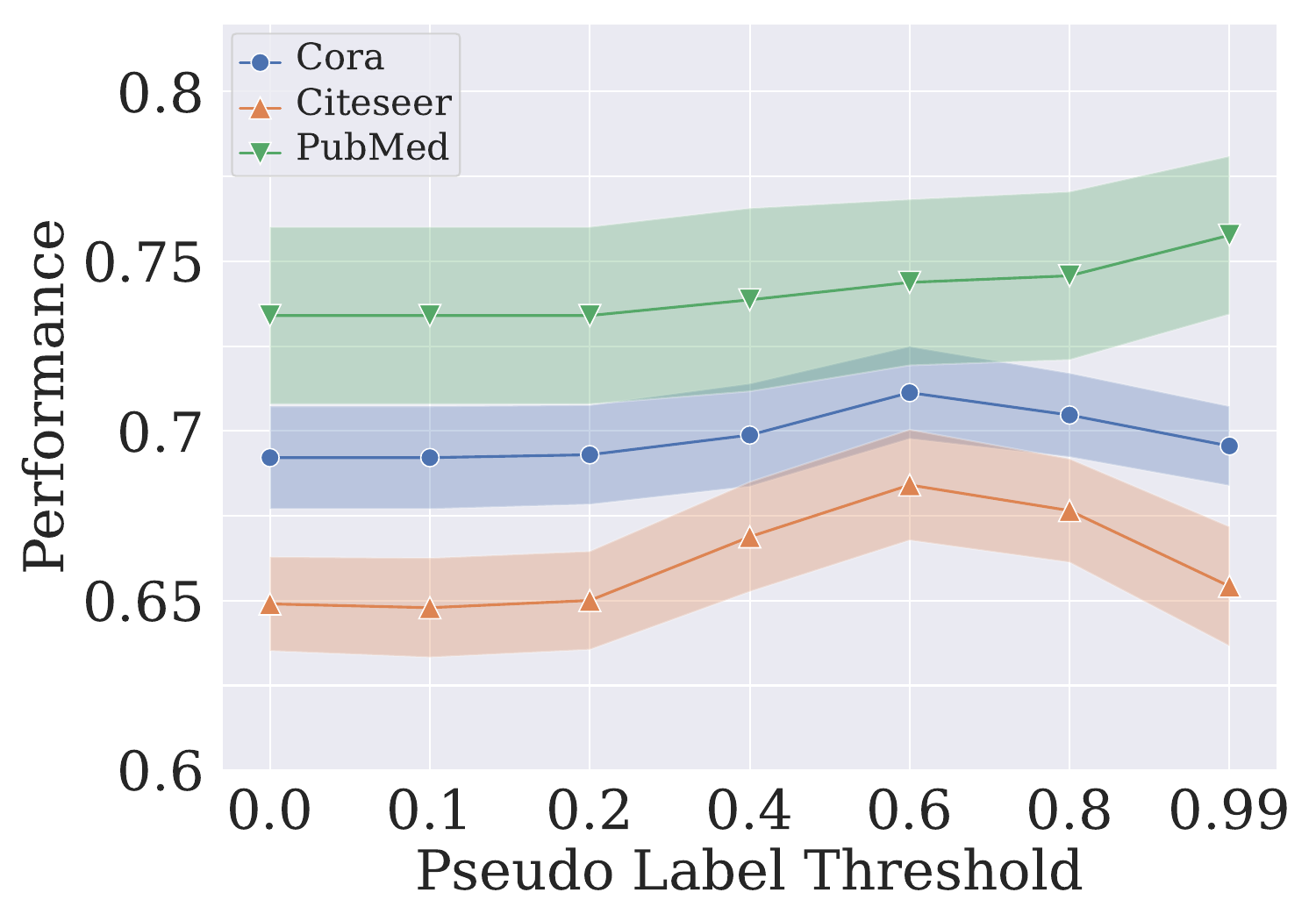}}}%
    \caption{\small{Parameter sensitivity analysis for $\kappa$ }}
    \label{fig:threshold_3_in_1}
    \vspace{-0.1cm}
\end{figure}

In this part, we investigate how the parameter $\kappa$ introduced in Section~\ref{sec:edgeclean} for controlling the filtering threshold of pseudo labels affects the final performance. A higher $\kappa$ means more pseudo labels will be filtered. In particular, we vary $\kappa$ from $0$ (without filtering out any pseudo labels) to $0.99$. The results are demonstrated in Figure~\ref{fig:threshold_3_in_1}. The figure reveals that with $\kappa = 0$, the model's performance is subpar across all three datasets. This outcome underlines the essentiality of filtering less confident pseudo labels for training the edge predictor. As $\kappa$ ascends, model performance generally improves due to the inclusion of more confident pseudo labels, thus providing more precise supervision. Upon further increasing $\kappa$, however, the model's performance begins to decline on the Cora and Citeseer datasets, attributed to the limited labeled data employed in training. Conversely, on PubMed, performance steadily escalates with $\kappa$ up to $0.99$. This pattern is predominantly owing to the relatively high confidence scores on PubMed, where over $52\%$ of pseudo labels possess a confidence score exceeding 0.99. Therefore, a meticulous tuning of $\kappa$ on PubMed may further improve the performance.

\section{Related Work}
\noindent{\bf Graph Active Learning (GAL).} AL has been studied specifically for GNNs. AGE \cite{cai2017active} mix multiple data selection metrics into its strategy. GPA \cite{hu2020graph} regards AL as a sequential decision process on graphs and trains a GNN-based policy network to learn the optimal query strategy. ANRMAB \cite{gao2018active} uses an active discriminative network representations with a multi-armed bandit mechanism for the GAL task. FeatProp \cite{wu2019active} selects nodes for labeling in the representations constructed by a parameter-free node feature propagation. LSCALE \cite{liulscale} exploits both labeled and unlabelled node representations for AL on graphs. ALG \cite{zhang2021alg} selects nodes by considering both node representativeness and informativeness and leverages decoupled GCNs to improve efficiency.  GRAIN \cite{zhang2021grain} performs data selection on graphs by achieving social influence maximization. RIM \cite{zhang2021rim} converts node selection to a social influence maximization problem and considers oracle noises. IGP \cite{zhang2022information} first proposes a soft-label approach to conduct AL for GNNs. ALLIE \cite{cui2022allie} designs AL specifically for large-scale imbalanced graph data. BIGENE \cite{zhang2022batch} proposes a multi-agent Q-network consisting of a GCN module and a gated recurrent unit module for data selection.  

\noindent{\bf Graph Structure Learning (GSL).} GSL aims to learn both a graph learning model and a graph structure simultaneously.  GCN-Jaccard \cite{wu2019adversarial} remove edges according to the Jaccard similarity of node features.  RGCN \cite{zhu2019robust} reduces the impacts of adversarial attacks by introducing variance-based attention weight in the message-passing. Pro-GNN \cite{jin2020graph} learns the graph structure and the GNN model simultaneously considering the low rank and sparsity property of clean graphs. RS-GNN \cite{dai2022towards} mines information in the noisy graph as an additional supervision signal to obtain a cleaned graph, which helps to improve predictions of GNNs. GEN \cite{wang2021graph} optimizes a graph structure model and an observation model to gain the optimal graph in an iterative manner. 
\section{Conclusion}\label{sec:conclusion}
Current GAL methods rely on the utilization of accurate graph information to select high-quality nodes for labeling. However, structural noise is ubiquitous in real-world graphs. We first investigate how the edge noise deteriorates the performance of widely used graph active learning models. After identifying the challenges of performing active learning on noisy graphs, we propose a novel iterative graph active learning framework by performing data selection and graph learning simultaneously. Not only high-quality data can be selected in \galp, but a cleaned graph will also be generated and utilized in the downstream graph tasks. Noticably, \galp enjoys a solid theoretical interpretation as an EM algorithm. Extensive experiments show that \galp has strong performance superiority over various baselines, especially when the existing noise is heavy. 

\appendix
\section{Experimental settings} 
Here, we provide detailed experimental settings such as the datasets, baselines, noise generation mechanism, and downstream GCN parameters to facilitate a better understanding of the Experiment Section in the main body of the paper.

\section{Dataset}
 \galp and baselines are evaluated on six datasets: Cora, Citeseer, Pubmed\cite{sen2008collective}, Amazon-photo, Amazon-Computer, and Coauthor-CS~\cite{shchur2018pitfalls}. The initial labeled set, $\mathcal{V}_{initial}$, is created by randomly sampling two nodes per class. We randomly select 50 nodes for the validation set, $\mathcal{V}_{valid}$, and 1000 nodes for the test set, $\mathcal{V}_{test}$. All methods select $8C$ nodes, $\mathcal{V}_{select}$, for labeling from the candidate pool, $\mathcal{V}_{pool}$, where $C$ represents the number of classes, and $\mathcal{V}_{pool}$ encompasses nodes not included in $\mathcal{V}_{initial}$, $\mathcal{V}_{valid}$, or $\mathcal{V}_{test}$. Ultimately, the labeled set $\mathcal{V}_{labeled} = \mathcal{V}_{initial} \cup \mathcal{V}_{select}$ is employed to train a downstream GCN model \cite{kipf2016semi}. We record the testing performance over $\mathcal{V}_{test}$. The experiments are performed 60 times using 10 different random initializations and six different random seeds. The report presents the average performance of the experiments.

\section{Baselines}
 We adopt the following classic and effective active learning models on graphs. (a) \AGE \cite{wu2019active} is a foundation active learning model on graphs, which combines several metrics including node centrality, node classification uncertainty, and node embedding representativeness to query nodes for labeling; (b) \LSCALE \cite{liulscale} is a graph active learning model that utilizes a self-supervised learning method to construct an informative embedding space for node selection; (c) \Grain \cite{zhang2021grain} considers the magnitude of the node influence and the diversity of the influence simultaneously as the criterion to select nodes for labeling; (d) \ALG \cite{zhang2021alg} leverages a model-free representativeness measurement and a cost-effective informativeness measurement empowered by a decoupled GCN model to conduct data selection; (e) \Random: We select nodes for labeling at random as the simplest strategy baseline.

\section{Noisy Graph Generation}
To demonstrate the robustness of the proposed model against the structural noise, we randomly link two unconnected nodes from different classes to introduce a noisy edge into the graph. The number of newly added noisy edges increases from 0\% of the number of edges (original graph) to 100\% (highly perturbed graph) by a stride of adding 20\% of the number of edges in the original graph. In addition, we leverage the recent state-of-the-art unsupervised attack model Contrastive Loss Gradient Attack (CLGA) \cite{zhang2022unsupervised} to generate attacked graph by adding 5\%, 10\%, 15\%, and 20\% additional noisy edges.

\section{Implementation Configuration}
Our framework is tuned on $\alpha$, $\beta$ and $ \kappa$. The batch size $S$ is 10 across all datasets. In assessing the GCN, we implement a two-layer GCN with a hidden dimension of 16 across all datasets, Amazon-Photo and Amazon-Computers being the exception. For Amazon-Photo and Amazon-Computers, we utilize a GCN model equipped with a hidden layer of 128 dimensions, as smaller hidden layers adversely impact the performance of the baselines.

\section{Time Complexity Analysis}
Here, we conducted a detailed time complexity analysis for \galp and it three key components to understand the scalability of the proposed method.
\noindent \textbf{Representation Model} We denote a graph as $\mathcal{G}=(\mathcal{V}, \mathcal{E})$, where $\mathcal{V}$ and $\mathcal{E}$ are the sets of nodes and edges, respectively. The time complexity for training the two-layer MLP per training epoch is $\mathcal{O}\left(\left|\mathcal{V}_l\right| D^2+\left|\mathcal{V}_l\right| D C\right)$, where $\left|\mathcal{V}_l\right|$ is the labeled node size, $D$ is the feature dimension and is also assumed as the hidden layer dimension. $C$ represents the number of classes. Since the cost for the graph constrastive loss is $\mathcal{O}\left(|\mathcal{V}|\left(S_{\text {pos }}+S_{\text {neg }}\right) D\right)$ ,where $S_{\text {pos }}$ and $S_{\text {neg }}$ are the positive and negative sample size respectively. In summary, the total complexity for training the representation model once can be expressed as follows:
$$
\mathcal{O}\left(\left|\mathcal{V}_l\right| D^2 P_{\text {rep}}+\left|\mathcal{V}_l\right| D C P_{\text {rep}}+|\mathcal{V}|\left(S_{\text {pos}}+S_{\text {neg}}\right) D P_{\text {rep}}\right)
$$
where $\left|\mathcal{V}_l\right|$ is the labeled nodes size and $P_{\text {rep }}$ is the number of epochs per training for the representation model.

\noindent \textbf{Edge-predictor} Following a similar analysis as before, the time complexity for training the edge-predictor and conducting inference is:
$$
\mathcal{O}\left(|\mathcal{V}| D^2 P_{e d g}+|\mathcal{E}| D\right)
$$
where $P_{\text {edg }}$ is the number of epochs per training for the edge-predictor.

\noindent \textbf{Node Selection} The time complexity for running K-means method is $\mathcal{O}(K D|\mathcal{V}| i)$, where $K$ is the number of cluster or the batch size of data selection, and $i$ is the number of iterations for K-means. Removing Well-Represented Nodes and calculating representativeness scores require $\mathcal{O}(K D|\mathcal{V}|)$, while the calculation of cleanliness score need the complexity of $\mathcal{O}(|\mathcal{E}| D)$. Therefore, in each batch selection, the time complexity for the node selection module in total is:
$$
\mathcal{O}(K D|\mathcal{V}| i+|\mathcal{E}| D)
$$
\noindent \textbf{Overall Framework} Let the total number of iterations for Phase 1 and Phase 2 be $I$. then the overall time complexity for the framework is:
\begin{align*}
\mathcal{O}(&I D(\left|\mathcal{V}_l\right| P_{\text {rep }}(D+C)+|\mathcal{V}|(S_{\text {pos }} + S_{\text {neg }}) P_{\text {rep }} \nonumber \\
& +D P_{\text {edg }}+K i)+|\mathcal{E}|)
\end{align*}

\section{Budget Sensitivity Analysis}
Labeling budget is a key factor in the active learning research. Here, we have conducted an additional study under 100\% random noise with varying budget sizes. The results are summarized in the Table~\ref{tab:budget}:

\begin{table}[H]
\centering
\caption{\small{ \textbf{Budget Sensitivity Analysis for \galp}}}
\begin{tabular}{ccccc}
\toprule
\textbf{Setting}   & \cora & \citeseer & \pubmed \\
\midrule
5 per class  &  66.16\%  & 60.84\%  & 69.30\% \\
8 per class  &  69.14\%  & 66.13\%  & 74.08\% \\
10 per class  &  70.39\%  & 67.65\%  & 75.76\% \\
15 per class  &  71.97\%  & 69.67\%  & 75.40\% \\
20 per class  &  72.68\%  & 70.38\%  & 75.74\% \\
\bottomrule
\end{tabular}
\label{tab:budget}
\end{table}
The result illustrates that increasing the labeling budget consistently enhances the the performance of our method on the Cora and Citeseer datasets. This underscores \galp's ability to continually improve the downstream GCN's performance even with the structural noises as the budget increases.

\section{Acknowledgement}
This research is supported by the National Science Foundation (NSF) under grant numbers  IIS-1909702, IIS-2153326 and IIS-2212145 and Army Research Office (ARO) under grant number W911NF21-1-0198.

\bibliographystyle{abbrv}
\bibliography{reference}

\newpage

\end{document}